\documentclass[10pt,twocolumn,letterpaper]{article}

\usepackage{cvpr}              

\definecolor{cvprblue}{rgb}{0.21,0.49,0.74}
\usepackage[pagebackref,breaklinks,colorlinks,allcolors=cvprblue]{hyperref}
\usepackage{stfloats}
\usepackage[most]{tcolorbox}

\def\paperID{650} 
\def\confName{CVPR}
\def\confYear{2025}

\title{GarmentPile: Point-Level Visual Affordance Guided Retrieval and Adaptation \\ for Cluttered Garments Manipulation}

\author{
Ruihai Wu\footnotemark[1] $^{1}$ \quad 
Ziyu Zhu\footnotemark[1] $^{2,1}$ \quad
Yuran Wang\footnotemark[1] $^{1}$ \quad
Yue Chen $^{1}$  \quad 
Jiarui Wang $^{1}$  \quad 
Hao Dong\footnotemark[2] $^{1}$ \quad  \\
$^1$CFCS, School of Computer Science, PKU \quad 
$^2$School of EECS, PKU
}

\begin{document}
\maketitle

\begin{abstract}
Cluttered garments manipulation poses significant challenges due to the complex, deformable nature of garments and intricate garment relations. Unlike single-garment manipulation, cluttered scenarios require managing complex garment entanglements and interactions, while maintaining garment cleanliness and manipulation stability. To address these demands, we propose to learn point-level affordance, the dense representation modeling the complex space and multi-modal manipulation candidates, while being aware of garment geometry, structure, and inter-object relations. Additionally, as it is difficult to directly retrieve a garment in some extremely entangled clutters, we introduce an adaptation module, guided by learned affordance, to reorganize highly-entangled garments into states plausible for manipulation. Our framework demonstrates effectiveness over environments featuring diverse garment types and pile configurations in both simulation and the real world. Project page: \href{https://garmentpile.github.io/}{https://garmentpile.github.io/}.
\end{abstract}
     
\section{Introduction}
\label{sec:intro}

\begin{figure}[htbp]
  \begin{center}
   \includegraphics[width=1.0\linewidth]{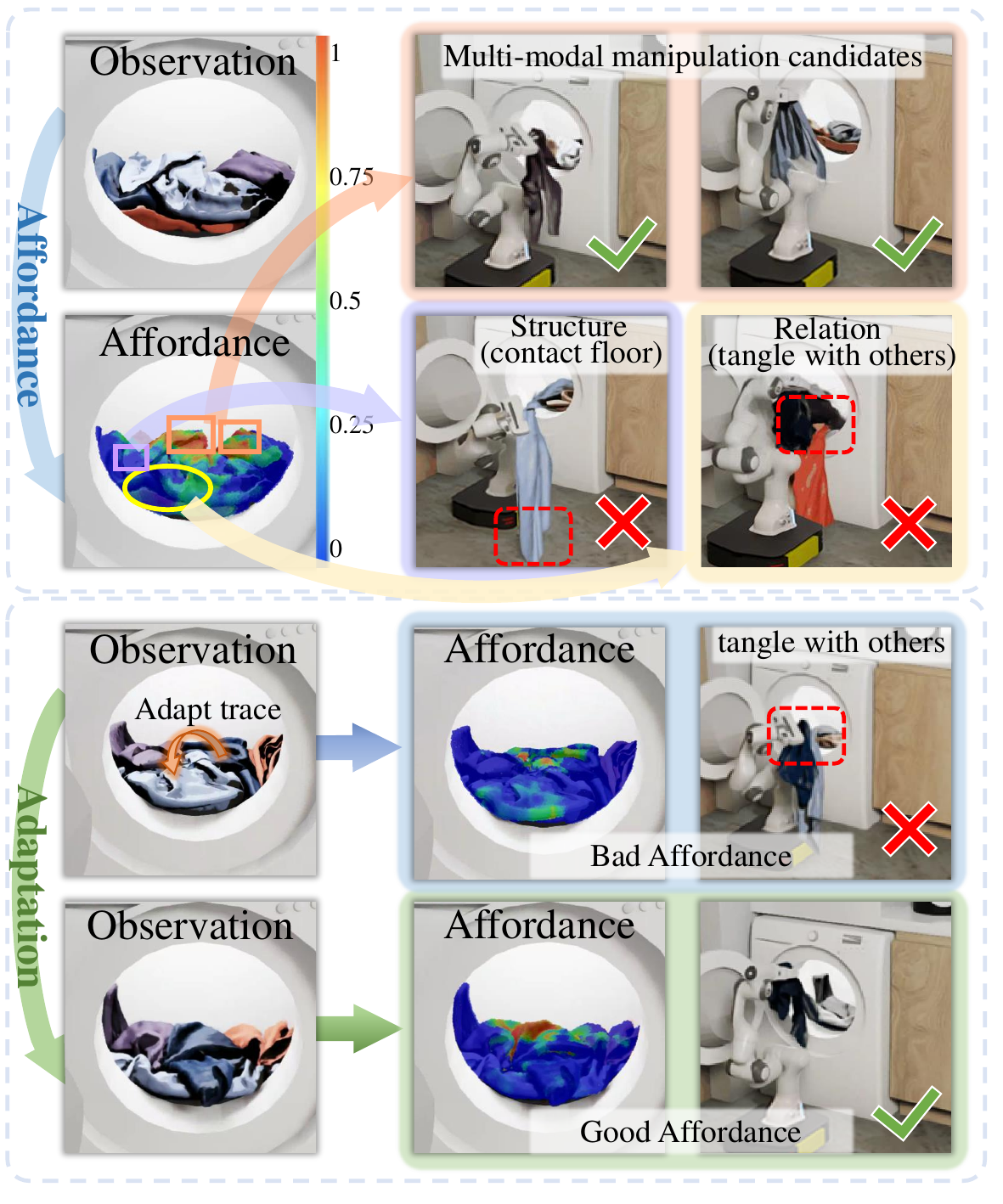}
  \end{center}
  \vspace{-5mm}\caption{\textbf{Point-Level Affordance for Cluttered Garments}. A higher score denotes the higher actionability for downstream retrieval.
  \textbf{\textit{Row 1}}: per-point affordance simultaneously reveals 2 garments suitable for retrieval.
  \textbf{\textit{Row 2}}: it is aware of garment structures (grasping edges leads other parts contacting floor) and relations (retrieving one garment while dragging nearby entangled garments out), and thus avoids manipulating on points leading to such failures.
  \textbf{\textit{Row 3} and \textit{4}}: 
  highly tangled garments may not have
  plausible manipulation points,
  affordance can guide reorganizing the scene, and thus garments plausible for manipulation will exist.
  }
\label{fig:teaser}
\end{figure}

Garments, such as shirts, dresses, and socks, are essential in daily life and pose significant challenges for human-assistive robots. Interacting with various types of garments to perform everyday tasks is crucial yet complex.
The complexity of garments comes from their high-dimensional state spaces, intricate kinematics, dynamics and diverse categories. Most studies focus on single-garment manipulation, such as unfolding~\cite{ha2022flingbot}, folding~\cite{avigal2022speedfolding, xue2023unifolding}, hanging~\cite{Wu_2024_CVPR}, and dressing up~\cite{Wang2023One}. However, many real-life scenarios involve multiple  cluttered garments, such as arranging clothes on a bed or retrieving items from a washing machine. In these cases, it is crucial to maintain cleanliness and avoid disturbing adjacent garments (failure cases in Figure~\ref{fig:teaser}).

Manipulating cluttered garments presents greater challenges than single-garment or cluttered rigid object manipulation. The more complex states in the clutters and the complicated interrelations between garments make it difficult to distinguish between different garments and infer their relations. 
Moreover, garment piles often involve multiple plausible retrieval garments (Figure~\ref{fig:teaser}, \textit{row 1}), further increasing the demands on the multi-modal representation capability of the learned manipulation policy.

Point-level affordance, derived from 3D point cloud input and representing the \textbf{per-point actionability} on the object for downstream tasks, is a suitable representation for cluttered garments manipulation.
First,
the per-point space supports representing complex states of cluttered scenes.
Also,
the per-point score can easily represent the multi-modal policy outputs (Figure~\ref{fig:teaser}, \textit{row 1}).
Most importantly,
the feature of each point is extracted from local to global,
capable of representing the garment local geometry information for grasping and retrieval, the structural information of each garment, and garment relation indicating whether the action would disturb other garments (Figure~\ref{fig:teaser}, \textit{row 2}).
For unseen garment clutters,
which easily exist as garments have nearly infinite deformations and combinations,
the above extracted  information (garment geometry, structure and relations) is consistent across scenes, making the representation easily generalize to novel scenarios.

However, affordance alone is not a universal solution. In extreme cases, such as highly tangled garments, the overall affordance is significantly reduced, meaning there may not exist manipulatable positions (Figure~\ref{fig:teaser}, \textit{row 3}), and thus need a person or robot to first reorganize the garments to a new state plausible for manipulation (Figure~\ref{fig:teaser}, from \textit{row 3} to \textit{row 4}).
Therefore, we further introduce a novel design, adaptation module, to mimic what people often do. By iteratively executing the pick-and-place actions, the adaptation module can use learned point-level affordance as the signal to efficiently reorganize cluttered garments.

The absence of suitable simulation environments also partly obstructs the research on cluttered garment manipulation. Previous works have primarily focused on the simulation and manipulation of single garments or simpler deformable objects~\cite{zhang2024adaptigraph, seita2021learning, wu2019learning, GPTFabric2024, wu2023learning,chen2023autobag,zhaole2024dexdlo}, rather than tackling the challenges posed by cluttered scenarios. To address this, we propose a new evaluation environment based on GarmentLab~\cite{lu2024garmentlab}, including 9 garment categories with various deformations and 3 representative scenarios: sofa, washing machine, and basket. 
Both qualitative and quantitative results from simulations and real-world experiments demonstrate the effectiveness of our framework.

In conclusion, our contributions mainly include:
\begin{itemize}
    \item We propose to study the novel task of cluttered garments manipulation, and build the pioneering environment with diverse scenarios covering different garment categories.
    \item We introduce point-level affordance learning for cluttered garments manipulation, with multiple novel designs to efficiently represent highly complex state and action spaces, and multi-modal policy outputs.
    \item We further develop the adaptation module guided by learned affordance, to efficiently adapt the cluttered garments to states easy to successfully manipulate.
    \item Extensive experiments in both simulation and real world demonstrate the effectiveness of our framework.
\end{itemize}

\section{Related Work}

\subsection{Visual Affordance for Robotic Manipulation}
\vspace{-0.5mm}
Visual affordance~\cite{gibson1977theory} indicates actionable possibilities on objects for various manipulation tasks. This approach has been widely used in grasping~\cite{corona2020ganhand, kokic2020learning, zeng2018robotic}, articulated manipulation~\cite{yuan2024general, kuang2024ram, act_the_part}, and scene interaction~\cite{interaction-exploration, ego-topo, hassanin2021visual, li2019putting}. Point-level affordance, in particular, assigns an actionability score to each point, and thus enabled fine-grained geometry understanding and improved cross-shape generalization in diverse tasks, such as articulated~\cite{mo2021where2act, wang2021adaafford}, bimanual~\cite{zhao2022dualafford} and deformable~\cite{Wu_2024_CVPR, wu2023learning} manipulation.
For cluttered garment scenarios, where garments often overlap and entangle, we empower point-level affordance with the awareness of garment structure and inter-relation, and further propose affordance-guided efficient scene adaptation.

\subsection{Garment Manipulation}
Among diverse categories of objects, such as rigid and articulated objects,
deformable objects pose particular challenges in representation and manipulation,
for the large state and action space,
as well as complex dynamics.
Manipulating 1D cables and ropes~\cite{zhang2024adaptigraph, seita2021learning, wu2019learning} and square-shaped fabrics~\cite{ganapathi2021learning, lee2020learning, wu2023learning, weng2022fabricflownet} have been relatively well-studied with vision, RL, or imitation based methods,
Garments featuring more diverse categories, shapes and deformations demonstrate more difficulty.
Current studies mainly focus on manipulating a single piece of garment, proposing methods for folding~\cite{xue2023unifolding, avigal2022speedfolding}, unfolding~\cite{ha2022flingbot, li2015regrasping, canberk2022cloth}, hanging~\cite{Wu_2024_CVPR, chen2023learning} garments,
and dressing up~\cite{Wang2023One, kotsovolis2024model, sun2024force}.
With the development of garment simulation environment~\cite{lu2024garmentlab},
we move a step towards building diverse scenarios with cluttered garments in various categories (\emph{e.g.}, shirts, dresses, gloves and socks),
and study the challenging cluttered garment manipulation.

\subsection{Cluttered Objects Manipulation}

Manipulating objects in cluttered environments is essential for tasks like grasping~\cite{sundermeyer2021contact, wang2021graspness, zeng2022robotic}, retrieving~\cite{xu2023joint, huang2021visual, kurenkov2020visuomotor}, and rearranging~\cite{goyal2022ifor, cheong2020relocate}. Cluttered scenes pose unique challenges, such as occlusions and complex spatial interactions, requiring advanced perception and control. Approaches using visual grounding~\cite{xu2023joint, lu2023vl}, object detection~\cite{li2023mobile}, and relational detection~\cite{li2024broadcasting} help locate and contextualize target objects, while end-to-end frameworks streamline pose prediction~\cite{breyer2021volumetric, fang2020graspnet}. Learned retrieval affordances~\cite{cheng2023learning, wang2021graspness, li2024broadcasting} further enhance adaptability in dense scenes. However, cluttered garments present unique challenges due to diverse shapes and deformations, complicating detection and relational understanding. We address this by proposing a point-level affordance approach with novel designs that capture the complex states and dynamics of garment piles, enabling more precise manipulation.
\begin{figure*}[t!]
  \begin{center}
   \includegraphics[width=1.0\linewidth]{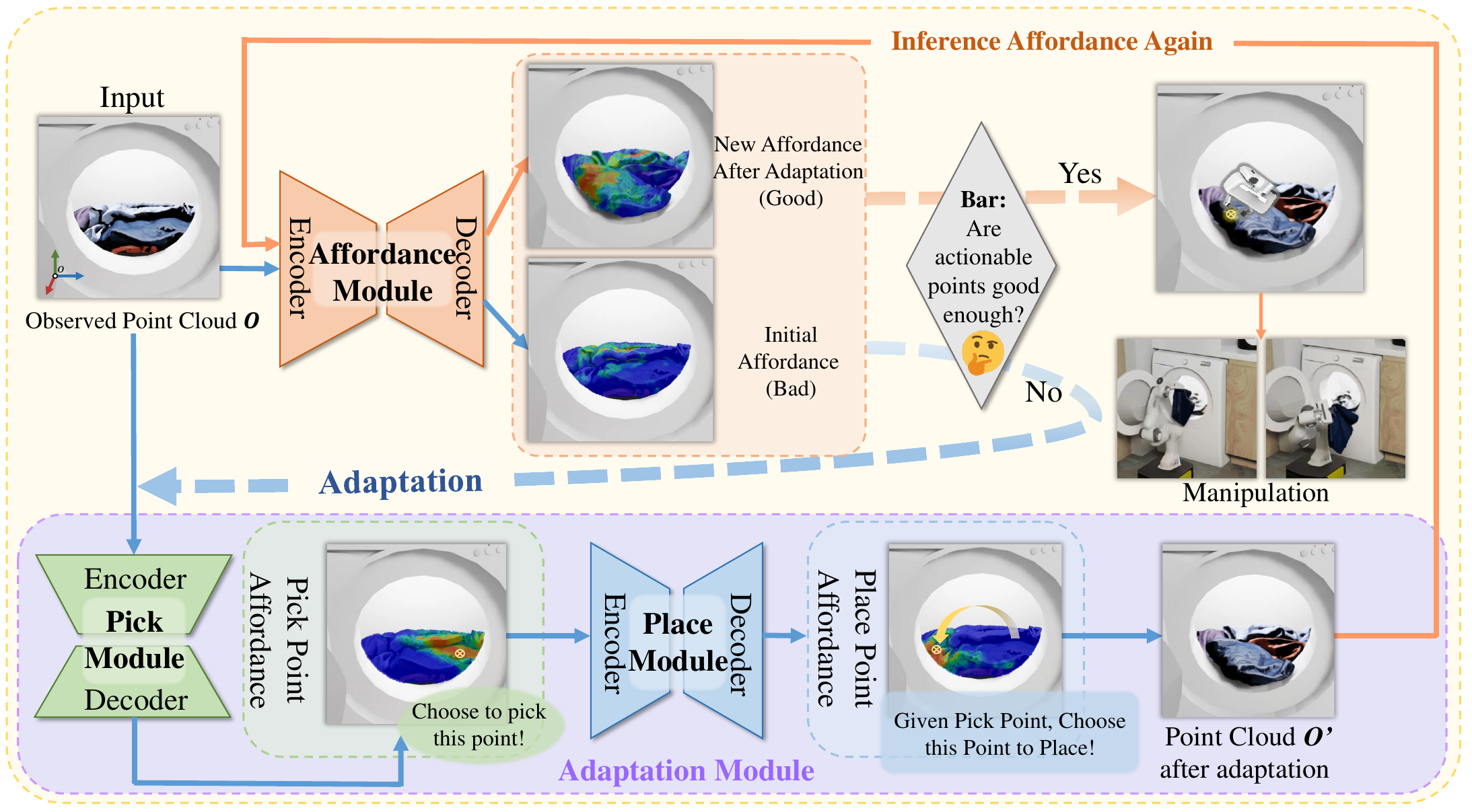}
  \end{center}
      \vspace{-2mm}
      \caption{\textbf{Framework Overview.} Given the observed point cloud, the Affordance Module predicts the initial point-level manipulation (retrieval)  affordance score. When actionability is not good enough, the framework proposes the adaptation pick-place action.
      It first predicts per-point pick affordance, and selects the pick point with the highest score, conditioned on which it predicts place affordance and selects the place point. After executing adaptation action, it receives a new point cloud and generates new affordance. When actionability is good enough, the robot retrieves on the point with the highest affordance score. 
      This loop is executed until all garments are retrieved.
      }
\label{fig:method}
\end{figure*}

\section{Problem Formulation}
Given a clutter of $k$ garments and its 3D point cloud observation $O \in \mathbb{R}^{N \times 3}$, we study garment retrieval, aiming to retrieve $k$ garments one-by-one while avoiding 2 common issues that may lead to uncleanliness or unsafety:
\begin{itemize}
    \item The target garment contacts the floor during the retrieval (Figure~\ref{fig:teaser}, \textit{row 2}, \textit{column 2}).
    \item When the retrieving one garment, others are dragged out (Figure~\ref{fig:teaser}, \textit{row 2}, \textit{column 3}).
\end{itemize}
We follow previous deformable manipulation studies~\cite{seita2021learning, wu2023learning, corl2020softgym} and use pick-and-place as action primitive. Since placing positions are usually predefined (e.g., placing garments into a basket or washing machine), we use the grasp point $p_{retrieve} \in \mathbb{R}^3$ with heuristic retrieval orientation as \textbf{retrieval action}. In case plausible retrieval garment is not available, we use pick-and-place action ($p_{pick} \in \mathbb{R}^3$ and $p_{place} \in \mathbb{R}^3$) as \textbf{adaptation action} to reorganize the scene.

We define point-level retrieval / pick / place affordance maps $A^{retrieve}$, $A^{pick}$, $A_{p_{pick}}^{place} \in \mathbb{R}^{N}$, each digit normalized to [0,1], indicating per-point actionability for retrieval / pick / place.
The point with highest score will be selected.
\section{Method}

\subsection{Overview}
We first describe directly learning point-level affordance for cluttered garment manipulation (Section ~\ref{method:aff}).
Then,
we describe learning the adaptation module (Section ~\ref{method:adaptation}),
by first learning point-level place affordance supervised by retrieval affordance (Section ~\ref{method:place}),
and point-level pick affordance supervised by place affordance (Section ~\ref{method:pick}).
Finally,
we introduce training strategy (Section ~\ref{method:strategy}).

\begin{figure*}[b]
  \begin{center}
   \includegraphics[width=1.0\linewidth]{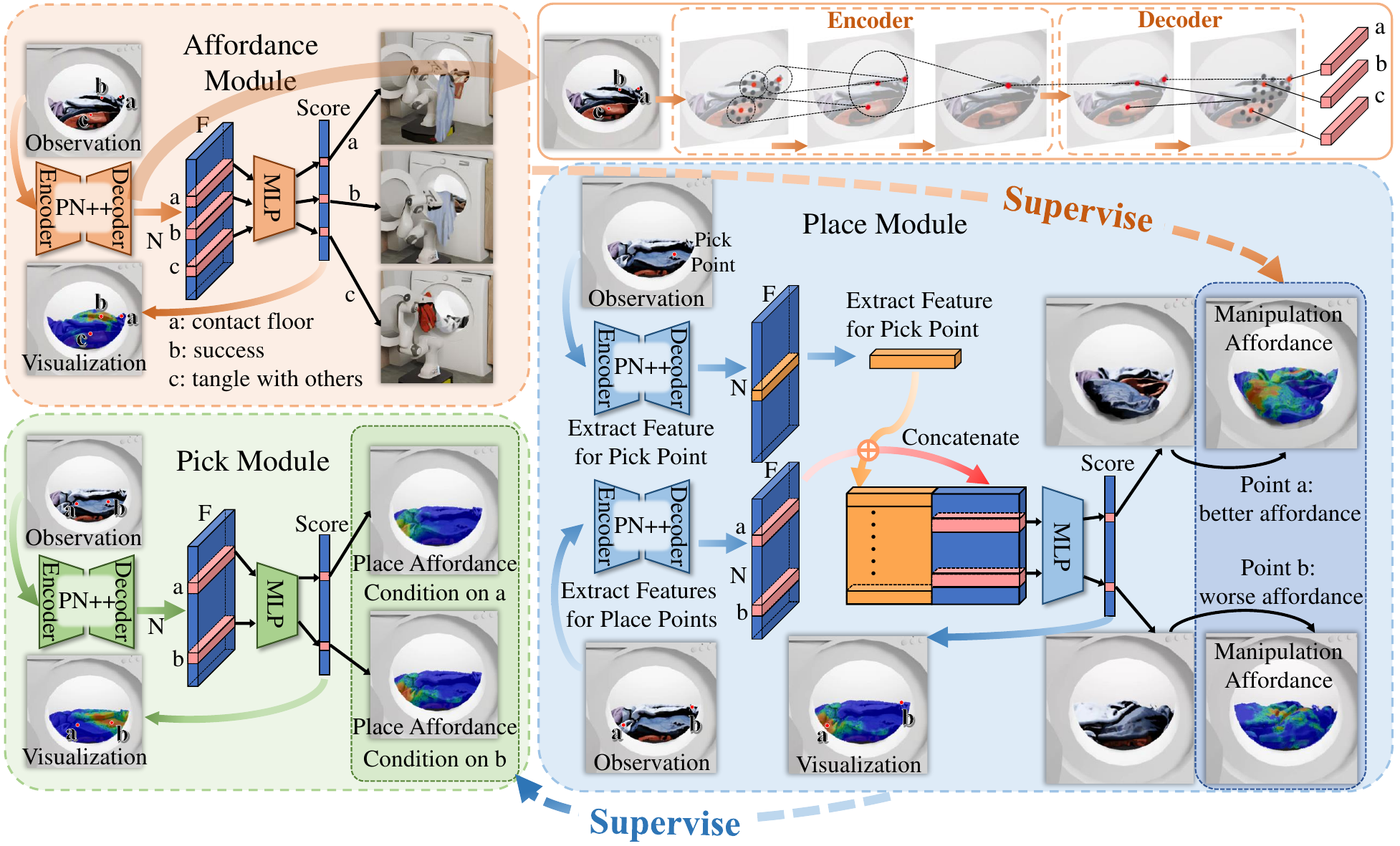}
  \end{center}
      \caption{\textbf{Learning Framework of Retrieval, Pick and Place Affordance.} Upper-left: the Affordance Module predicts the point-level (retrieval) affordance score for the downstream task. Upper-right: PointNet++ backbone aggregates both local and global features that facilitate incorporating garment geometry, structure and relation information for each point. Lower-right: the Place Module, which predicts the point-level place score conditioned on a pick point for adaptation, is supervised by the trained Affordance Module. Lower-left: the Pick Module, which predicts the point-level place score for adaptation, is supervised by the Place Module.}
\label{fig:learning}
\end{figure*}

\subsection{Point-Level Affordance for Retrieval}
\label{method:aff}

As described in Introduction and Problem Formulation,
the Retrieval Affordance Module (also denoted as Affordance Module for simplicity) $\mathcal{M}_{retrieve}$ predicts the per-point score map $\mathcal{A}^{retrieve}$ for each point.
Taking as input the point cloud observation $O$ of the garment clutter,
we extract the per-point feature using PointNet++~\cite{qi2017pointnet++} backbone feature extractor $\textbf{F}_{retrieve}$.
As demonstrated in Figure~\ref{fig:learning} (upper-right),
the per-point feature of PointNet++ aggregates the information of local geometry, global structure and garment relations,
each of which is essential for predicting whether the manipulation on the target point will succeed (affordance visualizations in ~\ref{fig:sequence} with analysis demonstrate such information in the learned affordance map).
For the point $p$, we get the feature $f^{retrieve}_p\in \mathbb{R}^{128}$, and parse it into Multi-Layer Perceptrons (MLPs) with sigmoid~\cite{DBLP:journals/corr/ElfwingUD17} activation function for normalization, we can get 1-dimension affordance prediction $\hat{g}^{retrieve}_p$ on $p$. 
We define the ground truth retrieval affordance score $g_{p}^{retrieve}$ on $p$ as $1$ (success) or $0$ (failure), by directly executing the retrieval action on $p$ and acquiring the manipulation result. 
We use Binary Cross Entropy (BCELoss) $L_{retrieve}$ to calculate the loss:

\begin{equation}
\begin{split}
    L_{retrieve} = & -\Big( g^{retrieve}_p \cdot \log(\hat{g}^{retrieve}_p) \\
                             & + (1 - g^{retrieve}_p) \cdot \log(1 - \hat{g}^{retrieve}_p) \Big)
\end{split}
  \label{eq:loss_retrieve}
\end{equation}

With trained $\mathcal{M}_{retrieve}$, given the 3D point cloud observation $O \in \mathbb{R}^{N\times3}$, we can first infer the point-level retrieval affordance map $A^{retrieval} \in \mathbb{R}^N$ and select the point $p_{retrieval}$ with the highest score for the retrieval action.

\subsection{Retrieval Affordance Guided Adaptation}
\label{method:adaptation}

Garments in clutters might be highly entangled, making it difficult to retrieve one garment without disturbing others in some situations,
where all points would have low affordance scores,
indicating that no point could be manipulated.
To deal with this situation, people often reorganize the garments (by picking-placing or stirring), until finding a plausible scene where the subsequent manipulation could be successful.
Therefore,
we mimic what people often do and propose the adaptation module by iteratively executing the pick-and-place actions to reorganize the scene.
The construction of the adaptation module depends on the learned affordance,
as it indicates whether the scene is plausible for manipulation.
To figure out the specific relationship between affordance and the success rate, we conducted extensive empirical tests and observed that the success rate is significantly high when \textbf{the portion of points with high retrieval affordance ($>0.9$)} (denoted as $P_{high}$) exceeds 0.1. Thus, we adapt multiple times until $P_{high}$ exceeds 0.1.

As pick-and-place composites a large action space, which is difficult for learning,
we disentangle each adaptation action into first predicting the pick point $p_{pick}$ and then the place point $p_{place}$ conditioned on $p_{pick}$.
As the state after placing can be estimated by the learned manipulation (retrieval) affordance,
we first learn given a specific pick point, the actionability for placing on each point, supervised by the retrieval affordance after the execution of ($p_{pick}$, $p_{place}$) action (Section~\ref{method:place}).
Then, with the learned place affordance for adaptation, we learn the affordance for $p_{pick}$ (Section~\ref{method:pick}).

\vspace{-2.5mm}
\subsubsection{Point-Level Place Affordance}
\label{method:place}
\vspace{-1mm}
The adaptation action is composed of a pick point $p_{pick}$ and a place point $p_{place}$.
With a pick point $p_{pick}$,
the Place Affordance Module $M_{place}$ rates the actionability of each point $p$ on whether placing $p_{pick}$ on $p$ will improve the scene.
The pick action $p_{pick}$ is difficult to directly get supervision, due to the diversity of the following place actions.
On the contrary, the place action $p$ conditioned on $p_{pick}$ can get the direct feedback from the adapted scene by checking the scene actionability (\emph{i.e.}, retrieval affordance) improvement.
Therefore, we first train $M_{place}$.

As demonstrated in Figure~\ref{fig:learning} (lower-right), for a target place point $p$, given as input the 3D point cloud $O$ and $p_{pick}$,
two PointNet++, $\textbf{F}^1_{place}$ and $\textbf{F}^2_{place}$,
respectively extracts the point feature $f^{place_1}_{p_{pick}}$ and $f^{place_2}_p$.
Then, their feature concatenation is parsed into MLPs to predict the place affordance $\hat{g}_{p|p_{pick}}^{place}$ normalized to [0, 1].
We execute the pick-and-place action from $p_{pick}$ to $p$ and get the new point cloud $O^\prime$, with the new affordance map. If the new affordance map exceeds the initial one by a margin, the ground truth place affordance ${g}_{p|p_{pick}}^{place}$ is set as $1$ otherwise $0$.
We use BCELoss $L_{place}$ to calculate the loss between ${g}_{p|p_{pick}}^{place}$ and $\hat{g}_{p|p_{pick}}^{place}$:
\begin{equation}
\begin{split}
    L_{place} = & - \Big( {g}_{p|p_{pick}}^{place} \cdot \log(\hat{g}_{p|p_{pick}}^{place}) \\
                             & + (1 - {g}_{p|p_{pick}}^{place}) \cdot \log(1 - \hat{g}_{p|p_{pick}}^{place}) \Big)
\end{split}
  \label{eq:loss_place}
\end{equation}

With the trained $\mathcal{M}_{place}$, given the 3D point cloud observation $O \in \mathbb{R}^{N\times3}$ and a specific $p_{pick}$, we can infer the point-level place affordance map $A^{place}_{p_{pick}} \in \mathbb{R}^N$ and select the point $p_{place}$ with the highest score for the place action.

\vspace{-2.5mm}
\subsubsection{Point-Level Pick Affordance}
\label{method:pick}
\vspace{-1mm}

For adaptation, with a stable Place Module that rates the actionability for each place point conditioned on any pick point,
we can further train the Pick Module that rates the actionability $g^{pick}_p$ for each point $p$,
supervised by the best following place action (\emph{i.e.}, the place point select by the learned Place Module) conditioned on pick $p$.

As demonstrated in Figure~\ref{fig:learning} (lower-left),
taking as input the point cloud observation $O$ of the garment clutter,
we extract the per-point feature using PointNet++ backbone feature extractor $\textbf{F}_{pick}$.
For the point $p$, we get the feature $f^{pick}_p\in \mathbb{R}^{128}$, and parse it into MLPs to predict the pick affordance $\hat{g}^{pick}_p$ normalized to [0, 1]. To get the groud truth score of this point, we use $\mathcal{M}_{place}$ to find the most suitable $p_{place}$ corresponding to $p$, and then execute the pick-and-place action from $p$ to $p_{place}$ to get the new point cloud $O^\prime$, with the new affordance map. If the new affordance map exceeds the initial one by a margin, the ground truth pick affordance $g^{pick}_p$ is set as $1$ otherwise $0$.

We use BCELoss $L_{pick}$ to calculate the loss between $g_{p}^{pick}$ and $\hat{g}^{pick}_p$:
\begin{equation}
\begin{split}
    L_{pick} = & -\Big( g_{p}^{pick} \cdot \log(\hat{g}_{p}^{pick}) \\
                             & + (1 - g_{p}^{pick}) \cdot \log(1 - \hat{g}_{p}^{pick}) \Big)
\end{split}
  \label{eq:loss_pick}
\end{equation}

With the trained $\mathcal{M}_{pick}$, given the 3D point cloud observation $O \in \mathbb{R}^{N\times3}$, we can infer the point-level pick affordance map $A^{pick} \in \mathbb{R}^N$ and select the point $p_{pick}$ with the highest score for the pick action.

\vspace{-1mm}
\subsection{Inference and Training Details}
\label{method:strategy}
\vspace{-1mm}

Figure~\ref{fig:method} (caption) describes \textbf{inference} pipeline and details.

Figure~\ref{fig:learning} (caption) describes the \textbf{training} pipeline.

For \textbf{training data, epoches and computing resources}, we use NVIDIA GeForce 4090 for training. We set batch size to be 128 to train (Retrieval) Affordance and Pick Affordance. While for Place Affordance, we set batch size to be 64 because there are two PointNet++ networks. We collect 20,000 pieces of data and train Retrieval Affordance for 120 epoches, as well as 8,000 pieces of data and train Pick and Place Affordance for 80 epoches. It takes fewer than 24 hours to train each module.

We further use \textbf{online data} to boost the robustness of during \textbf{training}.
Since cluttered garments have exceptionally diverse states,
the model trained on offline data might not work well in some unseen clutters.
Therefore,
based on the offline trained models, we collect the scenarios where the models do not well,
and further train the models on such data to improve the model robustness.
For example,
for the (Retrieval) Affordance Module,
we randomly sample scenes and use learned affordance model to infer the manipulation point $p_{retrieve}$.
If the manipulation failed after the actual execution,
we add this data into the buffer,
as this point is not actually suitable for manipulation.
When the buffer consists on 64 data,
we compose a batch of 128 data from the buffer and 64 offline data,
to train the model on the scenes in which it makes mistake predictions,
while maintaining the knowledge from the previous offline data.
We iterate this process as the sampled mistake distributions might have changes,
until the model shows consistent performance with low variance.
As we can acquire the manipulation or adaptation execution results for each module, 
the online adaptation proceeded for all modules.
\section{Experiments}
\vspace{-2mm}

\begin{figure*}[htbp]
  \begin{center}
   \includegraphics[width=1.0\linewidth]{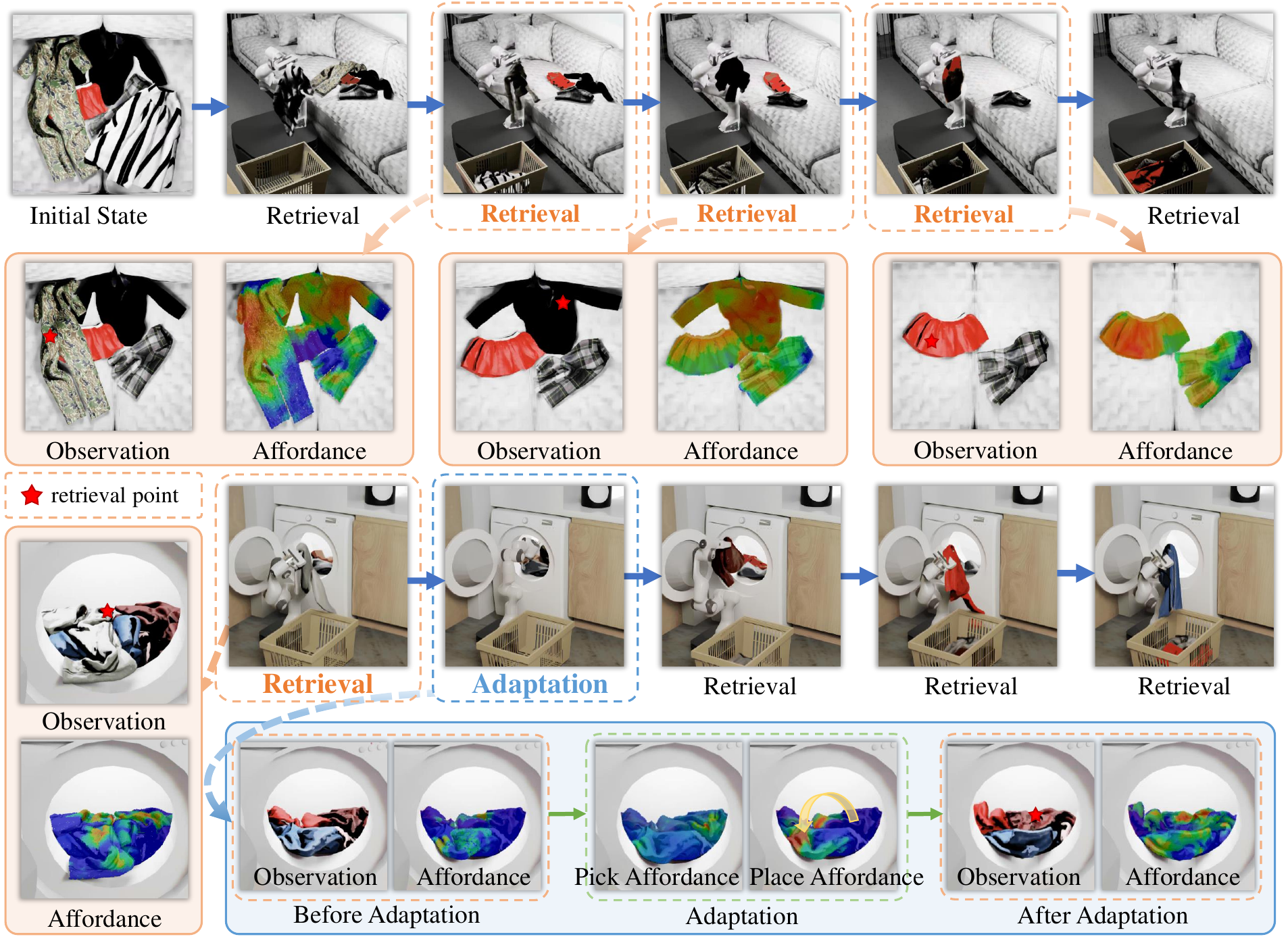}
  \end{center}
  \vspace{-5.5mm}\caption{\textbf{Example Manipulation Sequences} in WashingMachine and Sofa.}
  \vspace{-5mm}
\label{fig:sequence}
\end{figure*}

\subsection{Environment, Assets, Data and Evaluation}
\vspace{-1.5mm}

We use GarmentLab~\cite{lu2024garmentlab} built on Isaac Sim~\cite{Zhou_2024} as the environment, which supports simulating multi-material interaction coupling of garments.
We load 9 categories (dress, onesie, glove, hat, scarf, trousers, underpants, skirt and top) of 126 different garments from ClothesNet~\cite{zhou2023clothesnet}.
We construct 3 different types of representative and realistic scenes: 
\begin{itemize}
    \item \textbf{WashingMachine}  that contains piled garments and represents manipulation with spatial constraints.
    \item \textbf{Basket} that contains piled garments in the basket, representing manipulation with another spatial constraint type.
    \item \textbf{Sofa} that contains piled garments on the sofa, representing manipulation in the open space.
\end{itemize}
Each scene contains a maximum number of 5 garments in different types with various deformations. 
To obtain relatively complete point cloud, we place the sensor in front of washing machine for garments inside it, while above (top-down) garments on sofa and in baskets. Same poses are placed in the real world.
We use the success rate of manipulation for evaluation.

\subsection{Baselines and Ablations}

Due to the lack of direct studies in cluttered garments, we compare with methods on affordance learning, cluttered scene understanding and deformable object manipulation:
\begin{itemize}
    \item \textbf{Where2Act}~\cite{mo2021where2act} that predicts primitive-level (\emph{i.e.}, grasping or pulling) per-point actionability.
    \item \textbf{Support-M}, the modified version of \textbf{Supporting Relation}~\cite{kirillov2023segment} from cluttered rigid objects to cluttered garments, which first segments each object in the scene, and then retrieves the object not supported by others.
    \item \textbf{GPT-Fabric-M}, the modified \textbf{GPT-Fabric}~\cite{GPTFabric2024} to use GPT-4o~\cite{islam2024gpt} to infer spatial relations in garment clutters and retrieve the garment causing least influence to others.
\end{itemize}

\begin{figure*}[htbp]
  \begin{center}
   \includegraphics[width=1.0\linewidth]{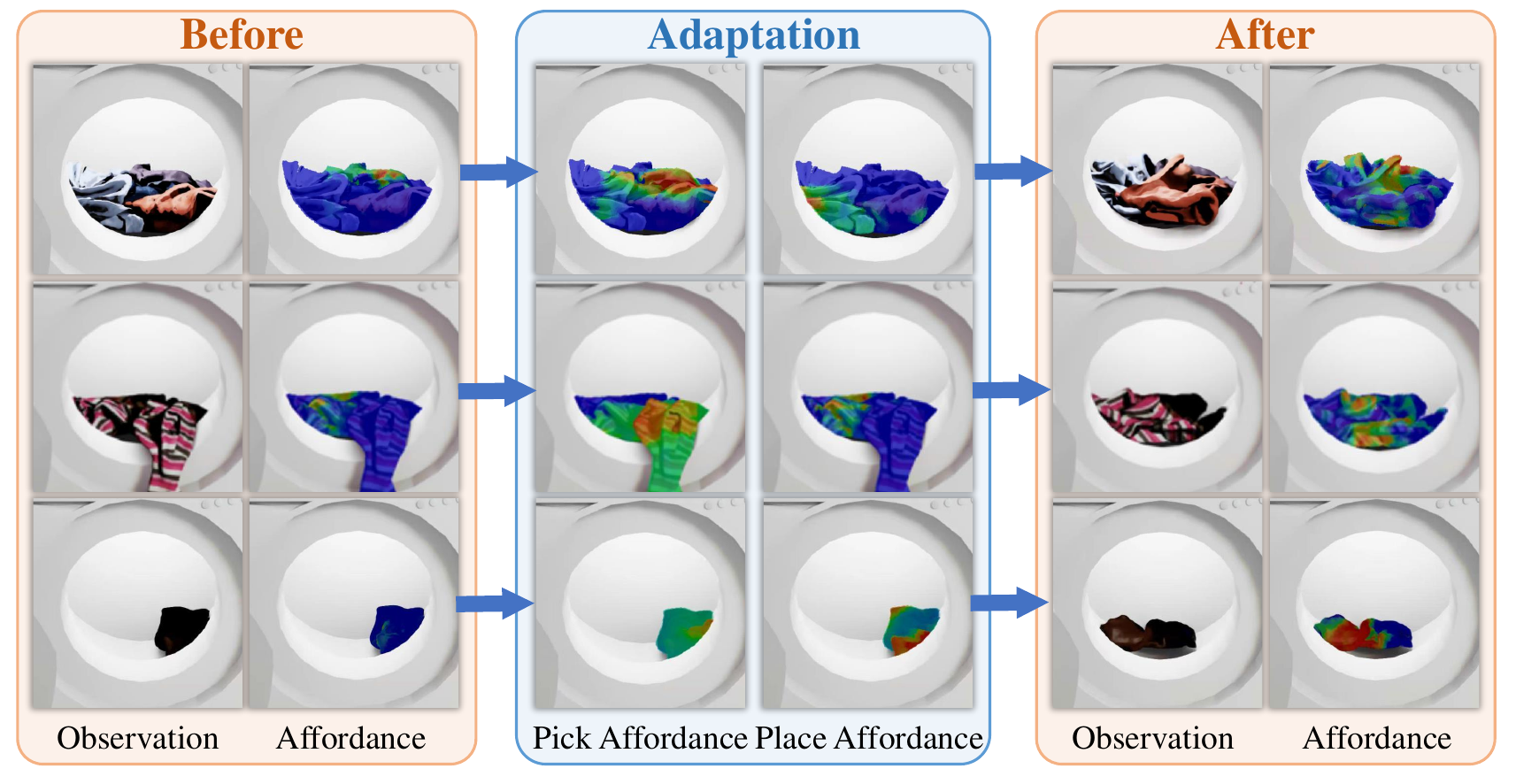}
  \end{center}
  \vspace{-5mm}\caption{\textbf{Retrieval Affordance before and after Adaptation, Adaptation Action indicated by Pick Affordance and Place Affordance}. When predicted retrieval affordance is not good enough(\textit{columns 1, 2}), the adaptation procedure will be triggered. The point with highest score in pick affordance will be chosen as $p_{pick}$ (\textit{column 3}) while the point with the highest score in place affordance will be chosen as $p_{place}$ (\textit{column 4}). After executing pick and place for adaptation, retrieval affordance has improved significantly (\textit{columns 5, 6}).}
\label{fig:adapt}
\end{figure*}

To demonstrate the necessity of the proposed adaptation module,
we compare with the following ablated versions:
\begin{itemize}
    \item \textbf{Ours w/o Adaptation} that removes the adaptation part.    \item \textbf{Ours w/o Pick Afford} that randomly selects the picking point instead of using trained pick affordance.
    \item \textbf{Ours w/o Place Afford} that randomly selects the placing point instead of using trained place affordance.
\end{itemize}

\subsection{Results and Analysis}
\label{sec:exp-analysis}

Figure~\ref{fig:sequence} demonstrates example manipulation trajectories  guided by affordance-based adaptation and manipulation.
Visualized affordance demonstrates the awareness of following representative features in cluttered garments:
\begin{itemize}
    \item Spots on garments where the manipulation will not cause garments to fall on the ground. In other words, the policy to some degree understands the overall structure of the garment, and thus will not grasp garment edges as other parts far from the grasping point may fall on the ground.
    \item Multi-modal retrieval points in clutters. Point-level representation is able and suitable to indicate diverse points in different garments plausible for retrieval.
    For example, in the Sofa scene, all upside garments have high scores.
    \item Geometries where the grasp operation will only grasp one garment instead of more. In other words, the junction between 2 garments do not have high affordance scores.
    \item Spots where the retrieval operation will not drag other garments as the side effect. For example, in the Sofa scene, only upside garments have high affordance scores.
\end{itemize}

\begin{table}
  \centering
  \begin{tabular}{@{}lccc@{}}
    \toprule
    Method & WashingMachine & Sofa & Basket \\
    \midrule
    Ours w/o Adaptation & 0.712 & 0.702 & 0.693 \\
    Ours w/o Pick Afford & 0.724 & 0.704 & 0.716 \\
    Ours w/o Place Afford & 0.778 & 0.743 & 0.731 \\
    Ours & \textbf{0.805} & \textbf{0.819} & \textbf{0.792} \\
    \bottomrule
  \end{tabular}
  \caption{\textbf{Ablation Study.} Section~\ref{sec:exp-analysis} provides detailed analysis.}
  \label{tab:ablation}
\end{table}

Figure \ref{fig:adapt} demonstrates the rationality and feasibility of our adaptation policy.  
What's more, from the procedure of adaptation we can also find out some evidence about how our model is aware of the structure of cluttered garment and take better methods about adapting the pile of garments:
\begin{itemize}
    \item \textit{Row 1}: the robot picks the red garment (supporting some upside garments) a little to the front, and thus this garment doe not support others.
    \item \textit{Row 2}: the sleeve leaks out of the washing machine and thus the garment might easily slip outside the machine when executing any retrieval action. The adaptation model picks the sleeve back into the washing machine.
    \item \textit{Row 3}: one small garment is hidden on the edge of the washing machine, and thus the robot cannot determine how many garments in the scene. The adaptation module choose to pick right of the front garment to the left, and the hidden garment is revealed.
\end{itemize}

Table~\ref{tab:ablation} shows results of ablation study. 
Pick affordance plays a crucial role as only limited points are plausible for picking for adaptation (\emph{i.e.}, picking most points do not improve the scene). Conditioned on a good pick point,
place affordance will further contribute to the better adaptation.

\begin{table}
  \centering
  \begin{tabular}{@{}lccc@{}}
    \toprule
    Method & WashingMachine & Sofa & Basket \\
    \midrule
    Where2Act~\cite{mo2021where2act} & 0.585 & 0.643 & 0.624 \\
    Support-M~\cite{kirillov2023segment} & 0.562 & 0.784 & 0.684 \\
    GPT-Fabric~\cite{GPTFabric2024} & 0.463 & 0.408 & 0.384 \\
    Ours & \textbf{0.805} & \textbf{0.819} & \textbf{0.792} \\
    \bottomrule
  \end{tabular}
  \caption{\textbf{Quantitative Comparisons with Baselines.} Our method outperforms baselines by a margin, with analyses in Section~\ref{sec:exp-analysis}.}
    \vspace{-5mm}
  \label{tab:baseline}
\end{table}

\begin{figure}[htbp]
  \begin{center}
   \includegraphics[width=1.0\linewidth]{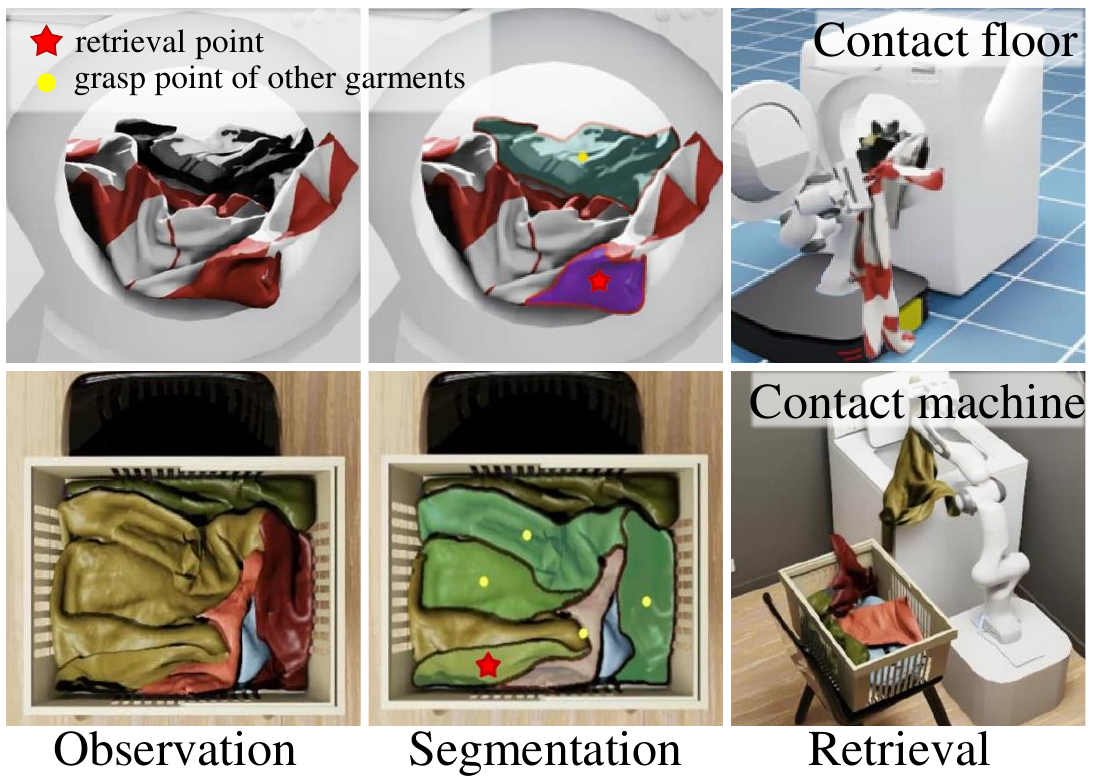}
  \end{center}
  \vspace{-6mm}\caption{\textbf{Manipulation Guided by Segmentation and Support Relation}. Foundation segmentation models cannot precisely segment garments in clutters, misleading the following manipulation.}
  \vspace{-7mm}
\label{fig:baseline_seg}
\end{figure}

\begin{figure}[htbp]
  \begin{center}
   \includegraphics[width=1.0\linewidth]{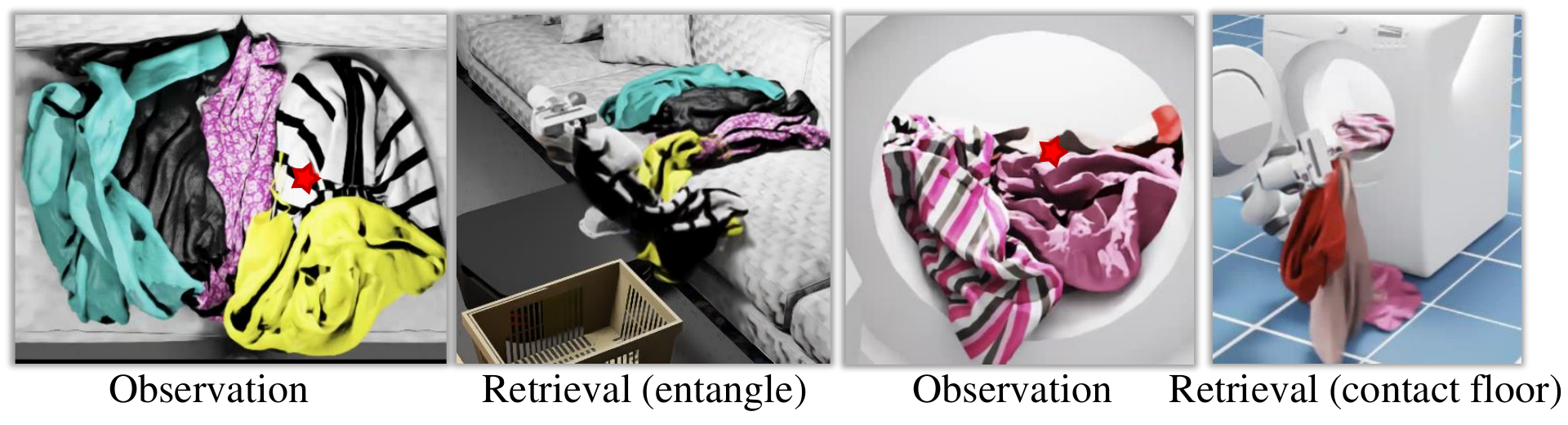}
  \end{center}
  \vspace{-5mm}
      \caption{\textbf{Manipulation Guided by GPT-4o.} GPT cannot precisely infer garment relations in complicated clutters.}
\label{fig:baseline_gpt}
\end{figure}

Table~\ref{tab:baseline} demonstrates the success rate of our method and baselines.
For \textbf{Where2Act}, as it only denotes affordance for pulling primitive, it cannot precisely guide the downstream retrieval task.
For \textbf{Support-M}, 
while it performs well in the Sofa scenario, its success rate drops sharply in WashingMachine and Basket. This is because garments on the sofa are in a relatively open space and thus show relatively intact shapes, making it not so difficult for foundation model to segment garments. 
However, WashingMachine and Basket scenarios involve severe occlusions and deformations, making garment segmentation significantly more challenging. As demonstrated in Figure~\ref{fig:baseline_seg}, foundation segmentation models mistakenly segments part of a garment as a whole garment. More visualizations are demonstrated in the supplementary material.
Besides, we observe that segmentation performance is influenced by lighting conditions and garment colors. For example, if the garments in a pile have similar colors or textures, chances are that they might not be segmented to separate individuals.
For \textbf{GPT-Fabric-M}, as shown in Figure~\ref{fig:baseline_gpt}, the high complexity of cluttered garments prevents GPT from accurately inferring garment relations. In the supplementary material, we provide the prompts and additional actions guided by GPT-4o. 
Notably, \textbf{Support-M} and \textbf{GPT-Fabric-M} require inference times of \textbf{3s} and \textbf{8s} respectively, which are significantly longer than affordance-based methods, making them less suitable for real-time decision-making and operation.

\vspace{-5mm}

\begin{figure}[htbp]
  \begin{center}
   \includegraphics[width=1.0\linewidth]{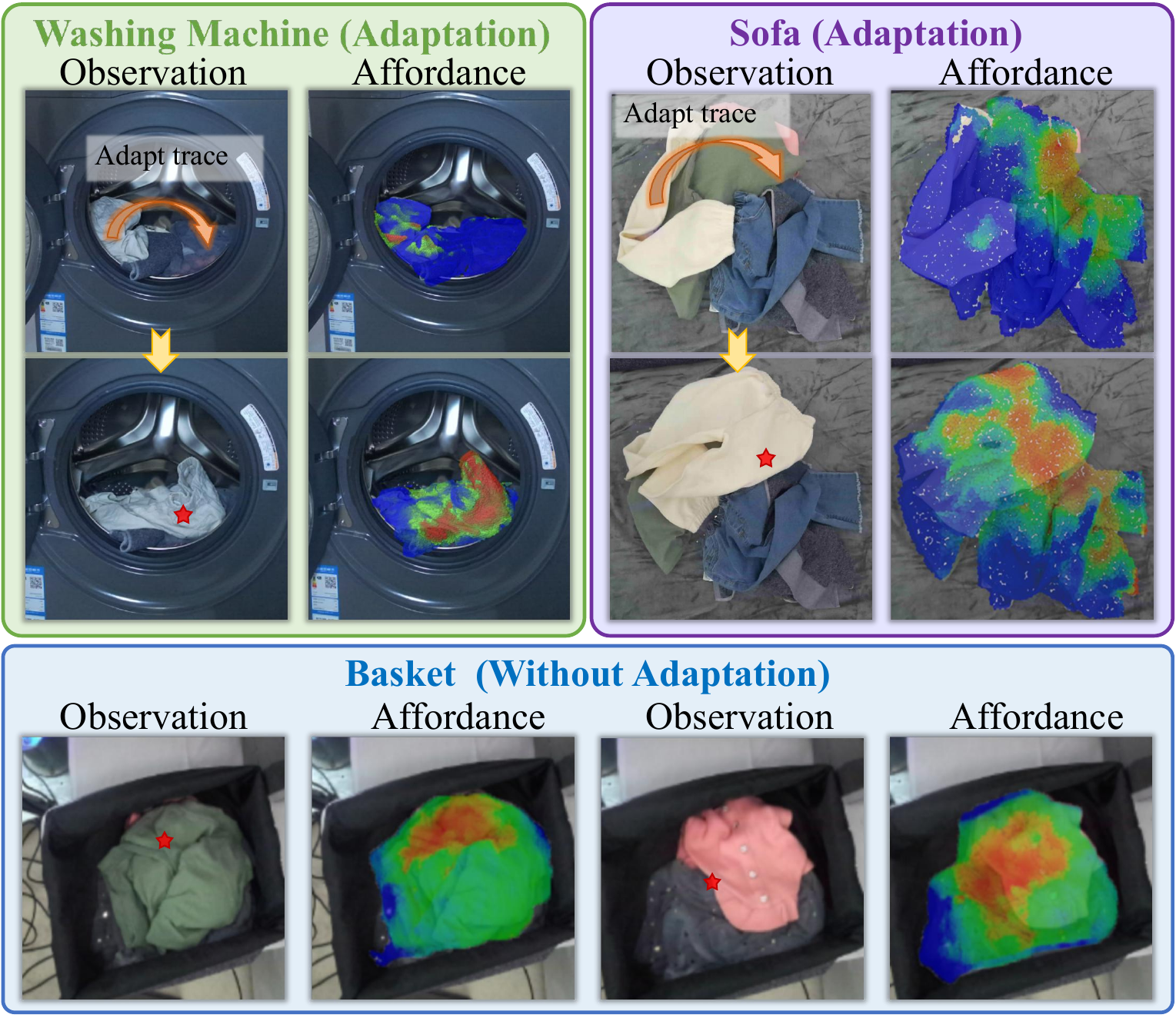}
  \end{center}
  \vspace{-5mm}
      \caption{\textbf{Real-World Results} of affordance and policy.}
\label{fig:real}
\vspace{-2mm}
\end{figure}

\vspace{-3.5mm}
\subsection{Real-World Evaluation}
\vspace{-1.5mm}

We use Franka Panda as the robot, and Kinect to capture depth images,
due to its slight noise problem demonstrated by previous studies in articulated object~\cite{ning2023where2explore} and garment manipulation~\cite{Wu_2024_CVPR}.
Table~\ref{tab:real} reports success rates.
Figure~\ref{fig:real} shows real-world observations, affordance and adaptation.
Supplementary video shows more demonstrations.

\begin{table}
  \centering
  \begin{tabular}{@{}lccc@{}}
    \toprule
    Method & WashingMachine & Sofa & Basket \\
    \midrule
    Where2Act~\cite{mo2021where2act} & 9 / 15 & 10 / 15 & 8 / 15 \\
    Support-M~\cite{kirillov2023segment} & 8 / 15 & 12 / 15 & 9 / 15 \\
    GPT-Fabric~\cite{GPTFabric2024} & 6 / 15 & 7 / 15 & 6 / 15 \\
    Ours & \textbf{12 / 15} & \textbf{13 / 15} & \textbf{12 / 15} \\
    \bottomrule
  \end{tabular}
  \vspace{-1.5mm}\caption{\textbf{Real-World Success Rate} in different scenarios.}
  \vspace{-5mm}
  \label{tab:real}
\end{table}

\section{Conclusion}

We propose point-level affordance, a dense representation, to capture complex cluttered garments scenarios and guide downstream manipulation.
Since cluttered garments may require initial adaptation before manipulation, we further introduce affordance-guided adaptation.
Experiments in diverse scenarios demonstrate the superiority of our method.


{
    \small
    \bibliographystyle{ieeenat_fullname}
    \bibliography{main}

\begin{thebibliography}{57}
\providecommand{\natexlab}[1]{#1}
\providecommand{\url}[1]{\texttt{#1}}
\expandafter\ifx\csname urlstyle\endcsname\relax
  \providecommand{\doi}[1]{doi: #1}\else
  \providecommand{\doi}{doi: \begingroup \urlstyle{rm}\Url}\fi

\bibitem[Avigal et~al.(2022)Avigal, Berscheid, Asfour, Kr{\"o}ger, and Goldberg]{avigal2022speedfolding}
Yahav Avigal, Lars Berscheid, Tamim Asfour, Torsten Kr{\"o}ger, and Ken Goldberg.
\newblock Speedfolding: Learning efficient bimanual folding of garments.
\newblock In \emph{IROS}, 2022.

\bibitem[Breyer et~al.(2021)Breyer, Chung, Ott, Siegwart, and Nieto]{breyer2021volumetric}
Michel Breyer, Jen~Jen Chung, Lionel Ott, Roland Siegwart, and Juan Nieto.
\newblock Volumetric grasping network: Real-time 6 dof grasp detection in clutter.
\newblock In \emph{Conference on Robot Learning}, pages 1602--1611. PMLR, 2021.

\bibitem[Canberk et~al.(2023)Canberk, Chi, Ha, Burchfiel, Cousineau, Feng, and Song]{canberk2022cloth}
Alper Canberk, Cheng Chi, Huy Ha, Benjamin Burchfiel, Eric Cousineau, Siyuan Feng, and Shuran Song.
\newblock Cloth funnels: Canonicalized-alignment for multi-purpose garment manipulation.
\newblock \emph{ICRA}, 2023.

\bibitem[Chen et~al.(2023{\natexlab{a}})Chen, Shi, Seita, Cheng, Kollar, Held, and Goldberg]{chen2023autobag}
Lawrence~Yunliang Chen, Baiyu Shi, Daniel Seita, Richard Cheng, Thomas Kollar, David Held, and Ken Goldberg.
\newblock Autobag: Learning to open plastic bags and insert objects, 2023{\natexlab{a}}.

\bibitem[Chen et~al.(2023{\natexlab{b}})Chen, Lee, Chappell, and Rojas]{chen2023learning}
Wei Chen, Dongmyoung Lee, Digby Chappell, and Nicolas Rojas.
\newblock Learning to grasp clothing structural regions for garment manipulation tasks.
\newblock In \emph{IROS}, 2023{\natexlab{b}}.

\bibitem[Cheong et~al.(2020)Cheong, Cho, Lee, Kim, and Nam]{cheong2020relocate}
Sang~Hun Cheong, Brian~Y Cho, Jinhwi Lee, ChangHwan Kim, and Changjoo Nam.
\newblock Where to relocate?: Object rearrangement inside cluttered and confined environments for robotic manipulation.
\newblock In \emph{ICRA}, 2020.

\bibitem[Corona et~al.(2020)Corona, Pumarola, Alenya, Moreno-Noguer, and Rogez]{corona2020ganhand}
Enric Corona, Albert Pumarola, Guillem Alenya, Francesc Moreno-Noguer, and Gr{\'e}gory Rogez.
\newblock Ganhand: Predicting human grasp affordances in multi-object scenes.
\newblock In \emph{Proceedings of the IEEE/CVF Conference on Computer Vision and Pattern Recognition}, pages 5031--5041, 2020.

\bibitem[Elfwing et~al.(2017)Elfwing, Uchibe, and Doya]{DBLP:journals/corr/ElfwingUD17}
Stefan Elfwing, Eiji Uchibe, and Kenji Doya.
\newblock Sigmoid-weighted linear units for neural network function approximation in reinforcement learning.
\newblock \emph{CoRR}, 2017.

\bibitem[Fang et~al.(2020)Fang, Wang, Gou, and Lu]{fang2020graspnet}
Hao-Shu Fang, Chenxi Wang, Minghao Gou, and Cewu Lu.
\newblock Graspnet-1billion: A large-scale benchmark for general object grasping.
\newblock In \emph{CVPR}, 2020.

\bibitem[Gadre et~al.(2021)Gadre, Ehsani, and Song]{act_the_part}
Samir~Yitzhak Gadre, Kiana Ehsani, and Shuran Song.
\newblock Act the part: Learning interaction strategies for articulated object part discovery.
\newblock In \emph{Proceedings of the IEEE/CVF International Conference on Computer Vision (ICCV)}, pages 15752--15761, 2021.

\bibitem[Ganapathi et~al.(2021)Ganapathi, Sundaresan, Thananjeyan, Balakrishna, Seita, Grannen, Hwang, Hoque, Gonzalez, Jamali, et~al.]{ganapathi2021learning}
Aditya Ganapathi, Priya Sundaresan, Brijen Thananjeyan, Ashwin Balakrishna, Daniel Seita, Jennifer Grannen, Minho Hwang, Ryan Hoque, Joseph~E Gonzalez, Nawid Jamali, et~al.
\newblock Learning dense visual correspondences in simulation to smooth and fold real fabrics.
\newblock In \emph{ICRA}, 2021.

\bibitem[Gibson(1977)]{gibson1977theory}
James~J Gibson.
\newblock The theory of affordances.
\newblock \emph{Hilldale, USA}, 1\penalty0 (2):\penalty0 67--82, 1977.

\bibitem[Goyal et~al.(2022)Goyal, Mousavian, Paxton, Chao, Okorn, Deng, and Fox]{goyal2022ifor}
Ankit Goyal, Arsalan Mousavian, Chris Paxton, Yu-Wei Chao, Brian Okorn, Jia Deng, and Dieter Fox.
\newblock Ifor: Iterative flow minimization for robotic object rearrangement.
\newblock In \emph{Proceedings of the IEEE/CVF Conference on Computer Vision and Pattern Recognition}, pages 14787--14797, 2022.

\bibitem[Ha and Song(2022)]{ha2022flingbot}
Huy Ha and Shuran Song.
\newblock Flingbot: The unreasonable effectiveness of dynamic manipulation for cloth unfolding.
\newblock In \emph{Conference on Robot Learning}. PMLR, 2022.

\bibitem[Hassanin et~al.(2021)Hassanin, Khan, and Tahtali]{hassanin2021visual}
Mohammed Hassanin, Salman Khan, and Murat Tahtali.
\newblock Visual affordance and function understanding: A survey.
\newblock \emph{ACM Computing Surveys (CSUR)}, 54\penalty0 (3):\penalty0 1--35, 2021.

\bibitem[Huang et~al.(2021)Huang, Han, Yu, and Boularias]{huang2021visual}
Baichuan Huang, Shuai~D Han, Jingjin Yu, and Abdeslam Boularias.
\newblock Visual foresight trees for object retrieval from clutter with nonprehensile rearrangement.
\newblock \emph{IEEE Robotics and Automation Letters}, 7\penalty0 (1):\penalty0 231--238, 2021.

\bibitem[Islam and Moushi(2024)]{islam2024gpt}
Raisa Islam and Owana~Marzia Moushi.
\newblock Gpt-4o: The cutting-edge advancement in multimodal llm.
\newblock \emph{Authorea Preprints}, 2024.

\bibitem[Kirillov et~al.(2023)Kirillov, Mintun, Ravi, Mao, Rolland, Gustafson, Xiao, Whitehead, Berg, Lo, et~al.]{kirillov2023segment}
Alexander Kirillov, Eric Mintun, Nikhila Ravi, Hanzi Mao, Chloe Rolland, Laura Gustafson, Tete Xiao, Spencer Whitehead, Alexander~C Berg, Wan-Yen Lo, et~al.
\newblock Segment anything.
\newblock \emph{arXiv preprint arXiv:2304.02643}, 2023.

\bibitem[Kokic et~al.(2020)Kokic, Kragic, and Bohg]{kokic2020learning}
Mia Kokic, Danica Kragic, and Jeannette Bohg.
\newblock Learning task-oriented grasping from human activity datasets.
\newblock \emph{IEEE Robotics and Automation Letters}, 5\penalty0 (2):\penalty0 3352--3359, 2020.

\bibitem[Kotsovolis and Demiris(2024)]{kotsovolis2024model}
Stelios Kotsovolis and Yiannis Demiris.
\newblock Model predictive control with graph dynamics for garment opening insertion during robot-assisted dressing.
\newblock In \emph{ICRA}, 2024.

\bibitem[Kuang et~al.(2024)Kuang, Ye, Geng, Mao, Deng, Guibas, Wang, and Wang]{kuang2024ram}
Yuxuan Kuang, Junjie Ye, Haoran Geng, Jiageng Mao, Congyue Deng, Leonidas Guibas, He Wang, and Yue Wang.
\newblock Ram: Retrieval-based affordance transfer for generalizable zero-shot robotic manipulation.
\newblock \emph{CoRL}, 2024.

\bibitem[Kurenkov et~al.(2020)Kurenkov, Taglic, Kulkarni, Dominguez-Kuhne, Garg, Mart{\'\i}n-Mart{\'\i}n, and Savarese]{kurenkov2020visuomotor}
Andrey Kurenkov, Joseph Taglic, Rohun Kulkarni, Marcus Dominguez-Kuhne, Animesh Garg, Roberto Mart{\'\i}n-Mart{\'\i}n, and Silvio Savarese.
\newblock Visuomotor mechanical search: Learning to retrieve target objects in clutter.
\newblock In \emph{2020 IEEE/RSJ International Conference on Intelligent Robots and Systems (IROS)}, pages 8408--8414. IEEE, 2020.

\bibitem[Lee et~al.(2020)Lee, Ward, Cosgun, Dasagi, Corke, and Leitner]{lee2020learning}
Robert Lee, Daniel Ward, Akansel Cosgun, Vibhavari Dasagi, Peter Corke, and Jurgen Leitner.
\newblock Learning arbitrary-goal fabric folding with one hour of real robot experience.
\newblock \emph{arXiv preprint arXiv:2010.03209}, 2020.

\bibitem[Li et~al.(2023)Li, Wei, Zhao, Yang, Li, and Zhang]{li2023mobile}
Dayou Li, Pengkun Wei, Chenkun Zhao, Shuo Yang, Yibin Li, and Wei Zhang.
\newblock A mobile manipulation system for automated replenishment in the field of unmanned retail.
\newblock In \emph{2023 IEEE International Conference on Mechatronics and Automation (ICMA)}, pages 644--649. IEEE, 2023.

\bibitem[Li et~al.(2019)Li, Liu, Kim, Wang, Yang, and Kautz]{li2019putting}
Xueting Li, Sifei Liu, Kihwan Kim, Xiaolong Wang, Ming-Hsuan Yang, and Jan Kautz.
\newblock Putting humans in a scene: Learning affordance in 3d indoor environments.
\newblock In \emph{CVPR}, 2019.

\bibitem[Li et~al.(2015)Li, Xu, Yue, Wang, Chang, Grinspun, and Allen]{li2015regrasping}
Yinxiao Li, Danfei Xu, Yonghao Yue, Yan Wang, Shih-Fu Chang, Eitan Grinspun, and Peter~K Allen.
\newblock Regrasping and unfolding of garments using predictive thin shell modeling.
\newblock In \emph{ICRA}, 2015.

\bibitem[Li et~al.(2024)Li, Wu, Lu, Ning, Shen, Zhan, and Dong]{li2024broadcasting}
Yitong Li, Ruihai Wu, Haoran Lu, Chuanruo Ning, Yan Shen, Guanqi Zhan, and Hao Dong.
\newblock Broadcasting support relations recursively from local dynamics for object retrieval in clutters.
\newblock In \emph{Robotics: Science and Systems}, 2024.

\bibitem[Lin et~al.(2020)Lin, Wang, Olkin, and Held]{corl2020softgym}
Xingyu Lin, Yufei Wang, Jake Olkin, and David Held.
\newblock Softgym: Benchmarking deep reinforcement learning for deformable object manipulation.
\newblock In \emph{CoRL}, 2020.

\bibitem[Lu et~al.(2024)Lu, Wu, Li, Li, Zhu, Ning, Shen, Luo, Chen, and Dong]{lu2024garmentlab}
Haoran Lu, Ruihai Wu, Yitong Li, Sijie Li, Ziyu Zhu, Chuanruo Ning, Yan Shen, Longzan Luo, Yuanpei Chen, and Hao Dong.
\newblock Garmentlab: A unified simulation and benchmark for garment manipulation.
\newblock In \emph{Advances in Neural Information Processing Systems}, 2024.

\bibitem[Lu et~al.(2023)Lu, Fan, Deng, Liu, Li, and Wang]{lu2023vl}
Yuhao Lu, Yixuan Fan, Beixing Deng, Fangfu Liu, Yali Li, and Shengjin Wang.
\newblock Vl-grasp: a 6-dof interactive grasp policy for language-oriented objects in cluttered indoor scenes.
\newblock In \emph{2023 IEEE/RSJ International Conference on Intelligent Robots and Systems (IROS)}, pages 976--983. IEEE, 2023.

\bibitem[Mo et~al.(2021)Mo, Guibas, Mukadam, Gupta, and Tulsiani]{mo2021where2act}
Kaichun Mo, Leonidas~J Guibas, Mustafa Mukadam, Abhinav Gupta, and Shubham Tulsiani.
\newblock Where2act: From pixels to actions for articulated 3d objects.
\newblock In \emph{CVPR}, 2021.

\bibitem[Nagarajan and Grauman(2020)]{interaction-exploration}
Tushar Nagarajan and Kristen Grauman.
\newblock Learning affordance landscapes for interaction exploration in 3d environments.
\newblock In \emph{NeurIPS}, 2020.

\bibitem[Nagarajan et~al.(2020)Nagarajan, Li, Feichtenhofer, and Grauman]{ego-topo}
Tushar Nagarajan, Yanghao Li, Christoph Feichtenhofer, and Kristen Grauman.
\newblock Ego-topo: Environment affordances from egocentric video.
\newblock In \emph{CVPR}, 2020.

\bibitem[Ning et~al.(2023)Ning, Wu, Lu, Mo, and Dong]{ning2023where2explore}
Chuanruo Ning, Ruihai Wu, Haoran Lu, Kaichun Mo, and Hao Dong.
\newblock Where2explore: Few-shot affordance learning for unseen novel categories of articulated objects.
\newblock \emph{arXiv preprint arXiv:2309.07473}, 2023.

\bibitem[Qi et~al.(2017)Qi, Yi, Su, and Guibas]{qi2017pointnet++}
Charles~Ruizhongtai Qi, Li Yi, Hao Su, and Leonidas~J Guibas.
\newblock Pointnet++: Deep hierarchical feature learning on point sets in a metric space.
\newblock \emph{NeurIPS}, 2017.

\bibitem[Raval et~al.(2024)Raval, Zhao, Zhang, Nikolaidis, and Seita]{GPTFabric2024}
Vedant Raval, Enyu Zhao, Hejia Zhang, Stefanos Nikolaidis, and Daniel Seita.
\newblock Gpt-fabric: Folding and smoothing fabric by leveraging pre-trained foundation models.
\newblock \emph{arXiv preprint arXiv:2406.09640}, 2024.

\bibitem[Seita et~al.(2021)Seita, Florence, Tompson, Coumans, Sindhwani, Goldberg, and Zeng]{seita2021learning}
Daniel Seita, Pete Florence, Jonathan Tompson, Erwin Coumans, Vikas Sindhwani, Ken Goldberg, and Andy Zeng.
\newblock Learning to rearrange deformable cables, fabrics, and bags with goal-conditioned transporter networks.
\newblock In \emph{ICRA}, 2021.

\bibitem[Sun et~al.(2024)Sun, Wang, Held, and Erickson]{sun2024force}
Zhanyi Sun, Yufei Wang, David Held, and Zackory Erickson.
\newblock Force-constrained visual policy: Safe robot-assisted dressing via multi-modal sensing.
\newblock \emph{RA-L}, 2024.

\bibitem[Sundermeyer et~al.(2021)Sundermeyer, Mousavian, Triebel, and Fox]{sundermeyer2021contact}
Martin Sundermeyer, Arsalan Mousavian, Rudolph Triebel, and Dieter Fox.
\newblock Contact-graspnet: Efficient 6-dof grasp generation in cluttered scenes.
\newblock In \emph{ICRA}, 2021.

\bibitem[Wang et~al.(2021)Wang, Fang, Gou, Fang, Gao, and Lu]{wang2021graspness}
Chenxi Wang, Hao-Shu Fang, Minghao Gou, Hongjie Fang, Jin Gao, and Cewu Lu.
\newblock Graspness discovery in clutters for fast and accurate grasp detection.
\newblock In \emph{Proceedings of the IEEE/CVF International Conference on Computer Vision}, pages 15964--15973, 2021.

\bibitem[Wang et~al.(2022)Wang, Wu, Mo, Ke, Fan, Guibas, and Dong]{wang2021adaafford}
Yian Wang, Ruihai Wu, Kaichun Mo, Jiaqi Ke, Qingnan Fan, Leonidas Guibas, and Hao Dong.
\newblock Adaafford: Learning to adapt manipulation affordance for 3d articulated objects via few-shot interactions.
\newblock \emph{European conference on computer vision (ECCV 2022)}, 2022.

\bibitem[Wang et~al.(2023)Wang, Sun, Erickson, and Held]{Wang2023One}
Yufei Wang, Zhanyi Sun, Zackory Erickson, and David Held.
\newblock One policy to dress them all: Learning to dress people with diverse poses and garments.
\newblock In \emph{RSS}, 2023.

\bibitem[Weng et~al.(2022)Weng, Bajracharya, Wang, Agrawal, and Held]{weng2022fabricflownet}
Thomas Weng, Sujay~Man Bajracharya, Yufei Wang, Khush Agrawal, and David Held.
\newblock Fabricflownet: Bimanual cloth manipulation with a flow-based policy.
\newblock In \emph{CoRL}, 2022.

\bibitem[Wu et~al.(2023{\natexlab{a}})Wu, Cheng, Zhao, Ning, Zhan, and Dong]{cheng2023learning}
Ruihai Wu, Kai Cheng, Yan Zhao, Chuanruo Ning, Guanqi Zhan, and Hao Dong.
\newblock Learning environment-aware affordance for 3d articulated object manipulation under occlusions.
\newblock In \emph{Thirty-seventh Conference on Neural Information Processing Systems}, 2023{\natexlab{a}}.

\bibitem[Wu et~al.(2023{\natexlab{b}})Wu, Ning, and Dong]{wu2023learning}
Ruihai Wu, Chuanruo Ning, and Hao Dong.
\newblock Learning foresightful dense visual affordance for deformable object manipulation.
\newblock In \emph{IEEE International Conference on Computer Vision (ICCV)}, 2023{\natexlab{b}}.

\bibitem[Wu et~al.(2024)Wu, Lu, Wang, Wang, and Dong]{Wu_2024_CVPR}
Ruihai Wu, Haoran Lu, Yiyan Wang, Yubo Wang, and Hao Dong.
\newblock Unigarmentmanip: A unified framework for category-level garment manipulation via dense visual correspondence.
\newblock In \emph{Proceedings of the IEEE/CVF Conference on Computer Vision and Pattern Recognition (CVPR)}, 2024.

\bibitem[Wu et~al.(2020)Wu, Yan, Kurutach, Pinto, and Abbeel]{wu2019learning}
Yilin Wu, Wilson Yan, Thanard Kurutach, Lerrel Pinto, and Pieter Abbeel.
\newblock Learning to manipulate deformable objects without demonstrations.
\newblock \emph{RSS}, 2020.

\bibitem[Xu et~al.(2023)Xu, Zhao, Zhou, Li, Pi, Zhu, Wang, and Xiong]{xu2023joint}
Kechun Xu, Shuqi Zhao, Zhongxiang Zhou, Zizhang Li, Huaijin Pi, Yifeng Zhu, Yue Wang, and Rong Xiong.
\newblock A joint modeling of vision-language-action for target-oriented grasping in clutter.
\newblock \emph{arXiv preprint arXiv:2302.12610}, 2023.

\bibitem[Xue et~al.(2023)Xue, Li, Xu, Li, Zheng, and Lu]{xue2023unifolding}
Han Xue, Yutong Li, Wenqiang Xu, Huanyu Li, Dongzhe Zheng, and Cewu Lu.
\newblock Unifolding: Towards sample-efficient, scalable, and generalizable robotic garment folding.
\newblock In \emph{CoRL}, 2023.

\bibitem[Yuan et~al.(2024)Yuan, Wen, Zhang, and Gao]{yuan2024general}
Chengbo Yuan, Chuan Wen, Tong Zhang, and Yang Gao.
\newblock General flow as foundation affordance for scalable robot learning.
\newblock \emph{arXiv preprint arXiv:2401.11439}, 2024.

\bibitem[Zeng et~al.(2018)Zeng, Song, Yu, Donlon, Hogan, Bauza, Ma, Taylor, Liu, Romo, et~al.]{zeng2018robotic}
Andy Zeng, Shuran Song, Kuan-Ting Yu, Elliott Donlon, Francois~R Hogan, Maria Bauza, Daolin Ma, Orion Taylor, Melody Liu, Eudald Romo, et~al.
\newblock Robotic pick-and-place of novel objects in clutter with multi-affordance grasping and cross-domain image matching.
\newblock In \emph{ICRA}, 2018.

\bibitem[Zeng et~al.(2022)Zeng, Song, Yu, Donlon, Hogan, Bauza, Ma, Taylor, Liu, Romo, et~al.]{zeng2022robotic}
Andy Zeng, Shuran Song, Kuan-Ting Yu, Elliott Donlon, Francois~R Hogan, Maria Bauza, Daolin Ma, Orion Taylor, Melody Liu, Eudald Romo, et~al.
\newblock Robotic pick-and-place of novel objects in clutter with multi-affordance grasping and cross-domain image matching.
\newblock \emph{IJRR}, 2022.

\bibitem[Zhang et~al.(2024)Zhang, Li, Hauser, and Li]{zhang2024adaptigraph}
Kaifeng Zhang, Baoyu Li, Kris Hauser, and Yunzhu Li.
\newblock Adaptigraph: Material-adaptive graph-based neural dynamics for robotic manipulation.
\newblock In \emph{RSS}, 2024.

\bibitem[Zhao et~al.(2023)Zhao, Wu, Chen, Zhang, Fan, Mo, and Dong]{zhao2022dualafford}
Yan Zhao, Ruihai Wu, Zhehuan Chen, Yourong Zhang, Qingnan Fan, Kaichun Mo, and Hao Dong.
\newblock Dualafford: Learning collaborative visual affordance for dual-gripper object manipulation.
\newblock \emph{ICLR}, 2023.

\bibitem[Zhaole et~al.(2024)Zhaole, Zhu, and Fisher]{zhaole2024dexdlo}
Sun Zhaole, Jihong Zhu, and Robert~B Fisher.
\newblock Dexdlo: Learning goal-conditioned dexterous policy for dynamic manipulation of deformable linear objects.
\newblock In \emph{2024 IEEE International Conference on Robotics and Automation (ICRA)}, pages 16009--16015. IEEE, 2024.

\bibitem[Zhou et~al.(2023)Zhou, Zhou, Liang, Yu, Zhao, Zeng, Lv, Luo, Wang, Yu, Chen, Lu, and Shao]{zhou2023clothesnet}
Bingyang Zhou, Haoyu Zhou, Tianhai Liang, Qiaojun Yu, Siheng Zhao, Yuwei Zeng, Jun Lv, Siyuan Luo, Qiancai Wang, Xinyuan Yu, Haonan Chen, Cewu Lu, and Lin Shao.
\newblock Clothesnet: An information-rich 3d garment model repository with simulated clothes environment.
\newblock \emph{ICCV}, 2023.

\bibitem[Zhou et~al.(2024)Zhou, Song, Xie, Shu, Ma, Liu, Yin, and See]{Zhou_2024}
Zhehua Zhou, Jiayang Song, Xuan Xie, Zhan Shu, Lei Ma, Dikai Liu, Jianxiong Yin, and Simon See.
\newblock Towards building ai-cps with nvidia isaac sim: An industrial benchmark and case study for robotics manipulation.
\newblock In \emph{Proceedings of the 46th International Conference on Software Engineering: Software Engineering in Practice}, page 263–274. ACM, 2024.

\end{thebibliography}
}

\clearpage
\appendix
\section{Results of Segment Anything}
\label{sec:seg}
\begin{figure}[htbp]
  \begin{center}
   \includegraphics[width=1.0\linewidth]{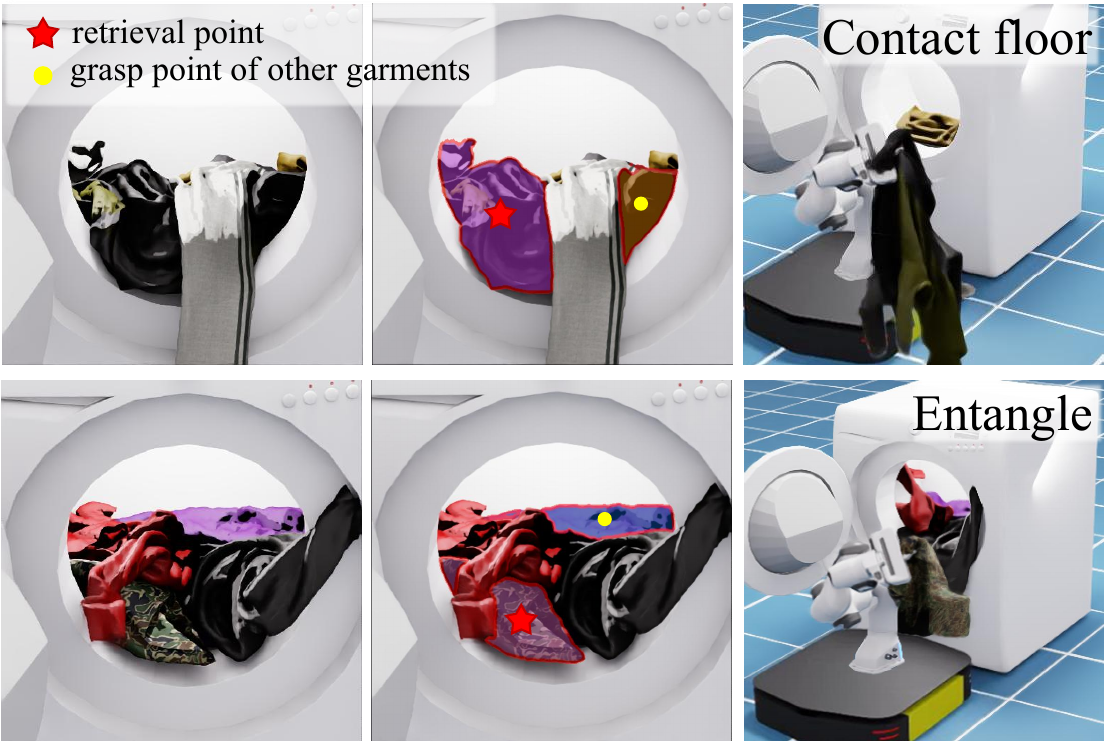}
  \end{center}
  \vspace{-5mm}\caption{Segment Anything Results in\textbf{ Washing Machine} Scene.}
  \vspace{-2mm}
\label{fig:wm_seg}
\end{figure}

\begin{figure}[htbp]
  \begin{center}
   \includegraphics[width=1.0\linewidth]{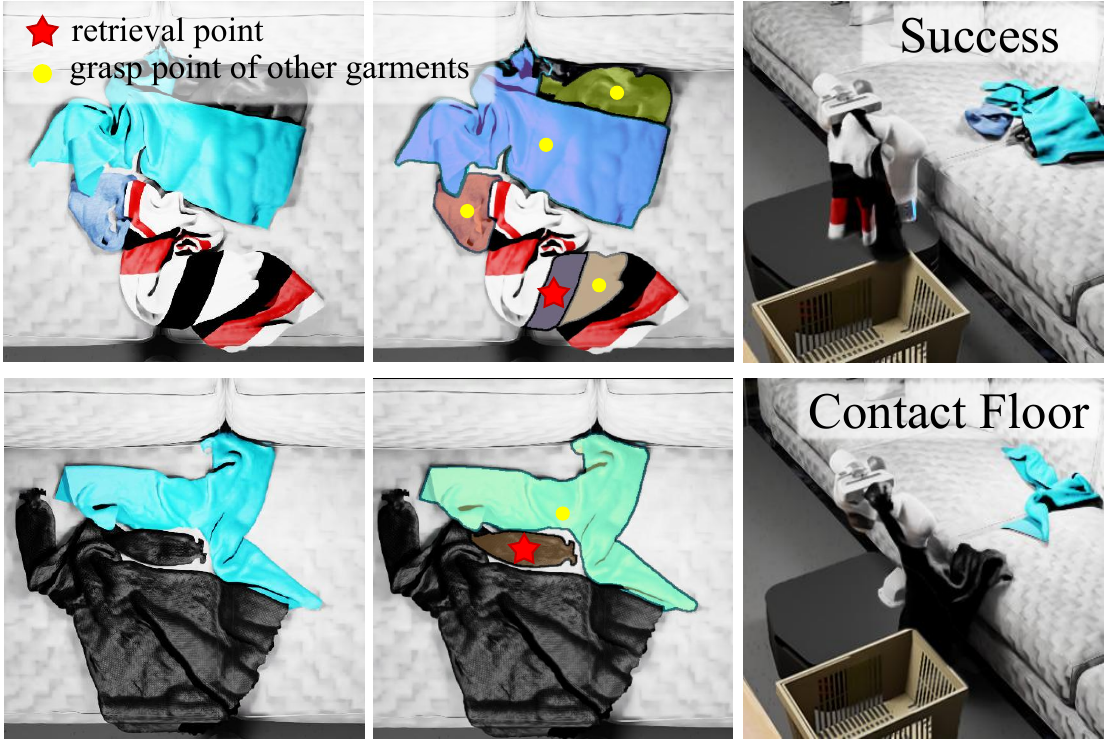}
  \end{center}
  \vspace{-5mm}\caption{Segment Anything Results in \textbf{Sofa} Scene.}
  \vspace{-2mm}
\label{fig:sofa_seg}
\end{figure}

\begin{figure}[htbp]
  \begin{center}
   \includegraphics[width=1.0\linewidth]{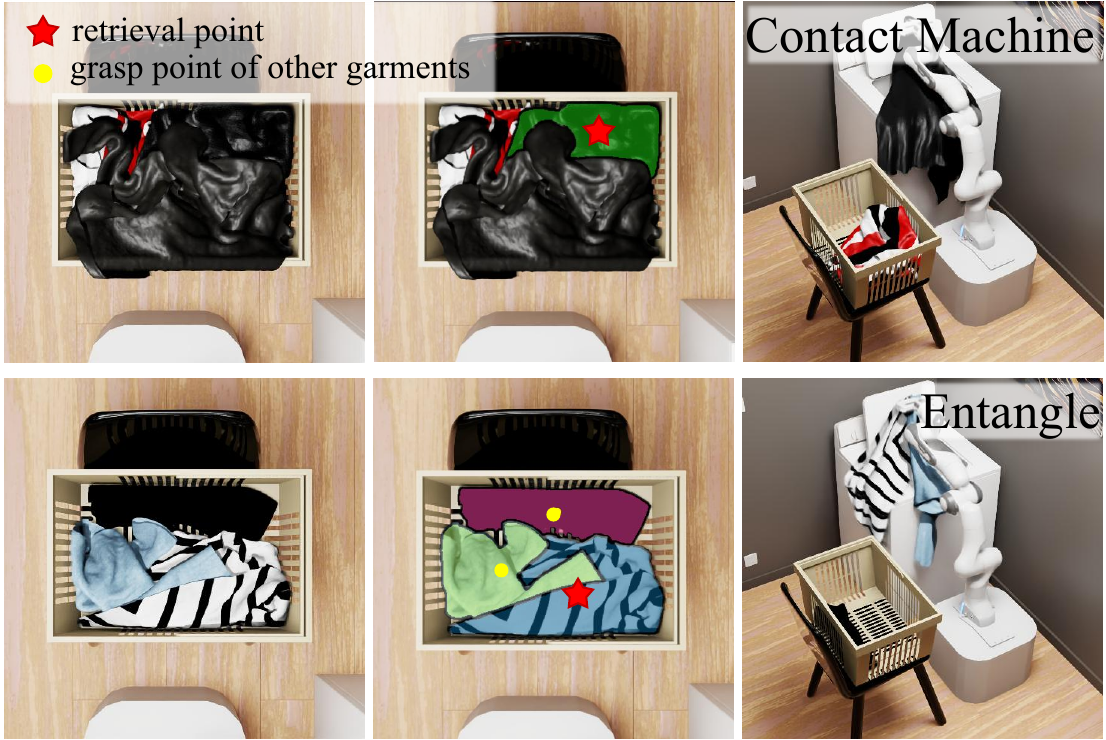}
  \end{center}
  \vspace{-5mm}\caption{Segment Anything Results in \textbf{Basket} Scene.}
  \vspace{-2mm}
\label{fig:basket_seg}
\end{figure}

We first explain how we use \textbf{Segment Anything} to obtain the manipulation point:

Given one RGB image, We first frame a region as the segmentation region of the Segment Anything model, which is the area where garment are concentrated. Based on the segmentation area, Segment Anything model will return us several segmentation part, then we get the coordinates of the center point of each part by calculating the average value of all points in the segmentation part, which is also called candidate grasp point (as shown by the yellow circle in the Figure~\ref{fig:wm_seg},~\ref{fig:sofa_seg} and~\ref{fig:basket_seg}). By comparing these candidate points, we choose the point closest to the exit (for washing machine) or with the highest height (for sofa and laundry basket) as the final retrieval point (as shown by the red star in the Figure~\ref{fig:wm_seg},~\ref{fig:sofa_seg} and~\ref{fig:basket_seg}).

We want to share some interesting circumstances in Segment Anything results: we found that the success rate of sofa scene is quite high based on Segment Anything, while the success rate of washing machine and basket scene based on Segment Anything is not so high (you can check the success rate in Table 2 of the main paper).

We think this is due to the characteristics of different scenes. In sofa scene, the stacking and occlusion relationship between garments is not so serious, so the model can segment the whole garment well and get the exact center point of garment, but for other scenes, the stacking and occlusion relationship between garments is too serious, which makes Segment Anything no longer perform well.

\section{Finetune SAM for Support-M}
\begin{figure}[htbp]
  \begin{center}
   \includegraphics[width=0.9\linewidth]{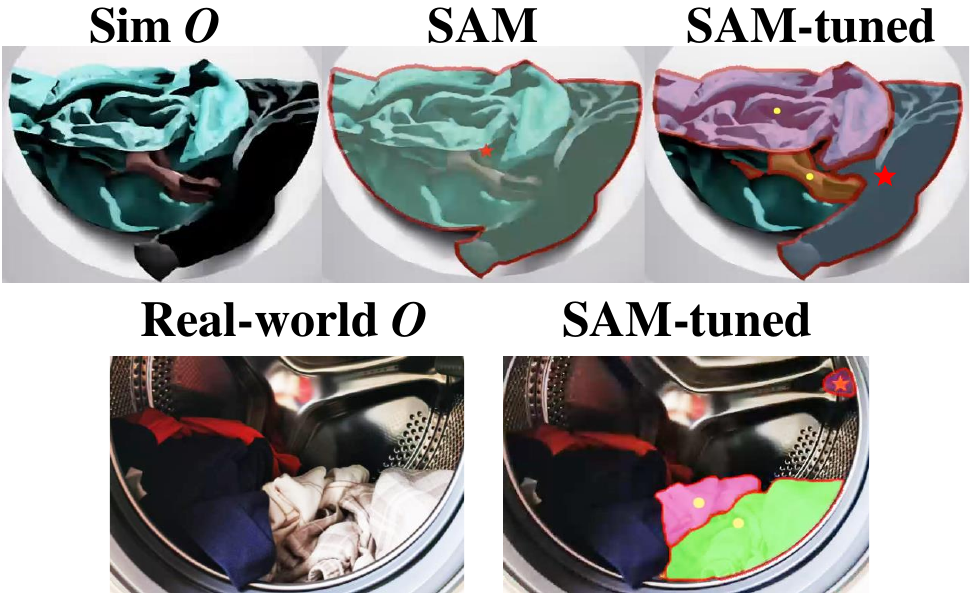}
  \end{center}
  \vspace{-5mm}\caption{\textbf{Segmentation of SAM} before/after finetuning. Segmentation is improved in simulation but not in the real world.}
  \vspace{-2mm}
\label{fig:sam}
\end{figure}

Due to the complexity of cluttered garments, it is difficult to obtain GT real-world segmentation masks, especially in real world scenarios.
We finetuned SAM using GT masks in simulation. Shown in Tab.~\ref{tab:sam}, baseline success rate improved $11\%$ in SIM (with \textbf{0.73} using GT segmentation as upper bound), while real world success rate remains \textbf{8 / 15} (qualitative results in Fig.~\ref{fig:sam}). 
Reasons: (1) only specific points, instead of all the segmented part, can be manipulated, which is only learned by our point-level representation; 
(2) gap between simulation and real-world images.

\begin{table}[!ht]
  \centering
  \scalebox{0.74}{
  \begin{tabular}{@{}lccccc|c@{}}
    \toprule
    Support-M    & SIM & SIM (tune) & SIM (GT seg) & Real & Real (tune) & Ours\\
    \midrule
    Succ Rate & 0.56 & 0.67 & 0.73 & 8/15 & 8/15 & 0.81\\

    \bottomrule
  \end{tabular}
  }
  \caption{\textbf{Success Rate of Support-M} with different segmentations in simulation and real world. \textit{GT seg} uses GT masks (upperbound).}
  \label{tab:sam}
\end{table}
\section{Results of Chatgpt-4o}
\label{sec:gpt}

We first explain how we use \textbf{Chatgpt-4o} to obtain the manipulation point:

Given one RGB image and one depth image, we first encode them in Base64 format and send them to Chatgpt-4o as conditions, while we also give Chatgpt-4o some relevant prompts to guide the model action, which is shown as below. Then Chatgpt-4o will return us one suitable retrieval point (if the point is not in the area of garments, we will make chatgpt-4o regenerate one point).

\begin{tcolorbox}[colframe=black, colback=pink!25, breakable, enhanced]
I will give you two images,\\
one is RGB and the other is a depth map.\\
The scene shows several pieces of clothing inside a basket. \textit{(this line should be changed according to different scenes)}\\
Assume you are a robot wanting to pick up each piece of clothing from the basket individually, \textit{(this line should be changed according to different scenes)}\\
ensuring that no other garments are accidentally pulled out during the process.\\
Provide me with the optimal grabbing point as precisely as possible,\\
which should be the coordinates of a pixel in the RGB image.\\
The point you select must meet the basic requirements of being located on a piece of clothing.\\
After generating a point, check if it is on the clothing. If not, select a new point and repeat the process until it is on the clothing.\\
Note, you only need to return the precise coordinates of the pixel you consider optimal. And precision is very important.\\
No additional information is required.\\
For example: (201, 313)
\end{tcolorbox}

\begin{figure}[htbp]
  \begin{center}
   \includegraphics[width=1.0\linewidth]{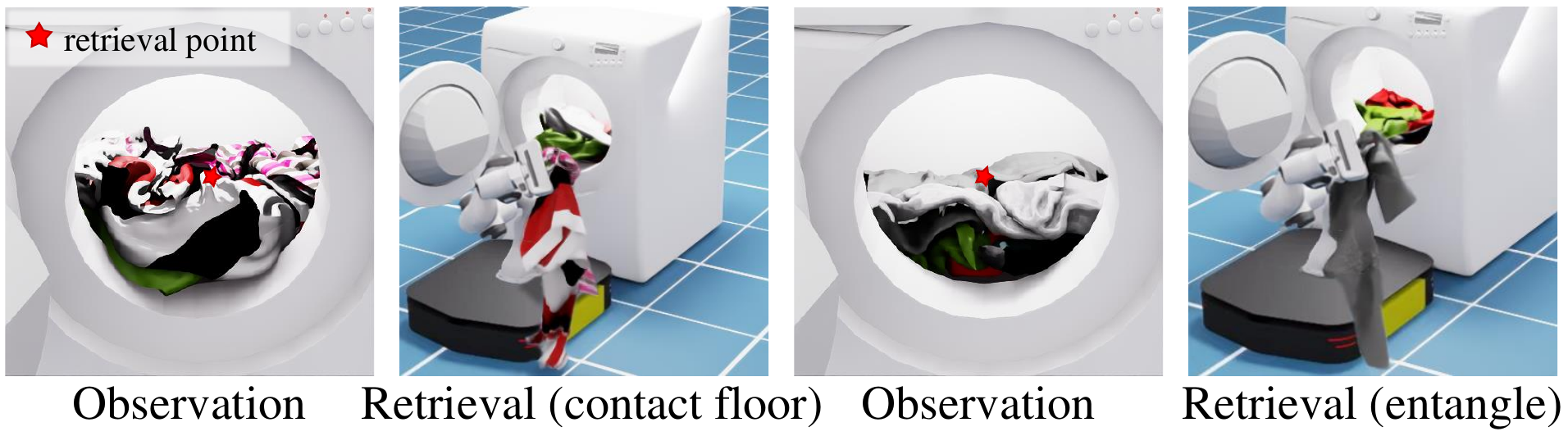}
  \end{center}
  \vspace{-5mm}\caption{Chatgpt-4o Results in \textbf{Washing Machine} Scene.}
  \vspace{-2mm}
\label{fig:gpt_wm}
\end{figure}

\begin{figure}[htbp]
  \begin{center}
   \includegraphics[width=1.0\linewidth]{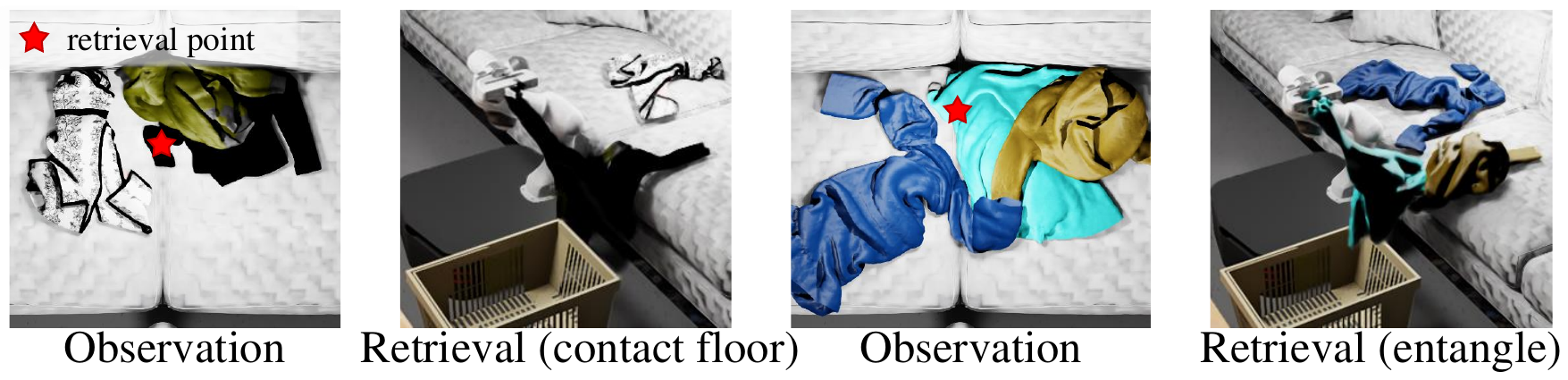}
  \end{center}
  \vspace{-5mm}\caption{Chatgpt-4o Results in \textbf{Sofa} Scene.}
  \vspace{-2mm}
\label{fig:gpt_sofa}
\end{figure}

\begin{figure}[htbp]
  \begin{center}
   \includegraphics[width=1.0\linewidth]{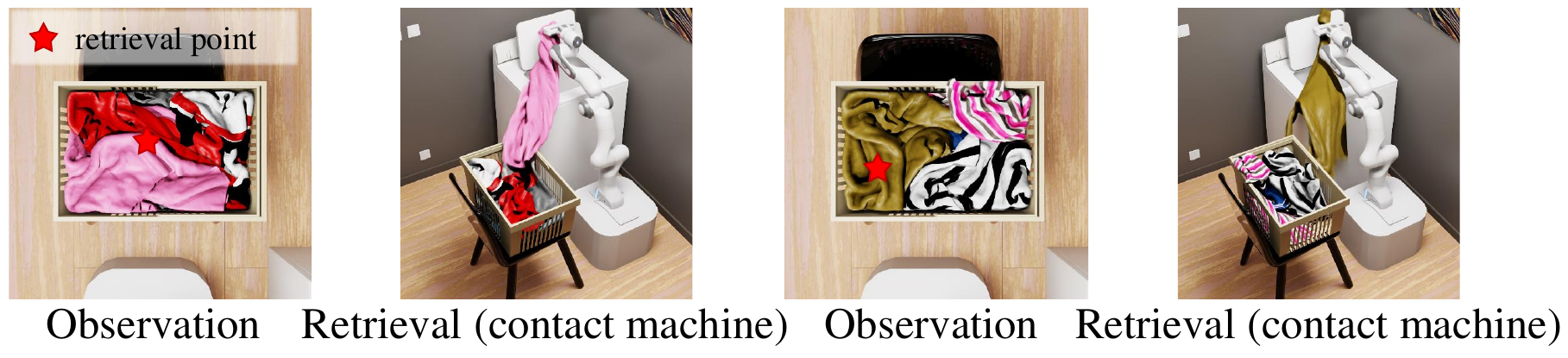}
  \end{center}
  \vspace{-5mm}\caption{Chatgpt-4o Results in\textbf{ Basket} Scene.}
  \vspace{-2mm}
\label{fig:gpt_basket}
\end{figure}

Here we show some additional results about Chatgpt-4o in the scene of washing machine (Figure~\ref{fig:gpt_wm}), sofa (Figure~\ref{fig:gpt_sofa}) and laundry basket (Figure~\ref{fig:gpt_basket}).

We unfortunately find that the performance of Chatgpt-4o is far below expectations. The model seems to just select random points, which were unreasonable in most cases and unsuitable for effective manipulation.
\section{Results of Real Machine}
\label{sec:real_machine}

In this part we show the whole procedure about retrieval and adaptation in our real machine scenes (including washing machine, sofa and laundry basket). It is worth mentioning that our model can work well without fine-tuning based on online data in the real world, which proves that our model has good generalization and robustness.

\begin{figure}[htbp]
  \begin{center}
   \includegraphics[width=1.0\linewidth]{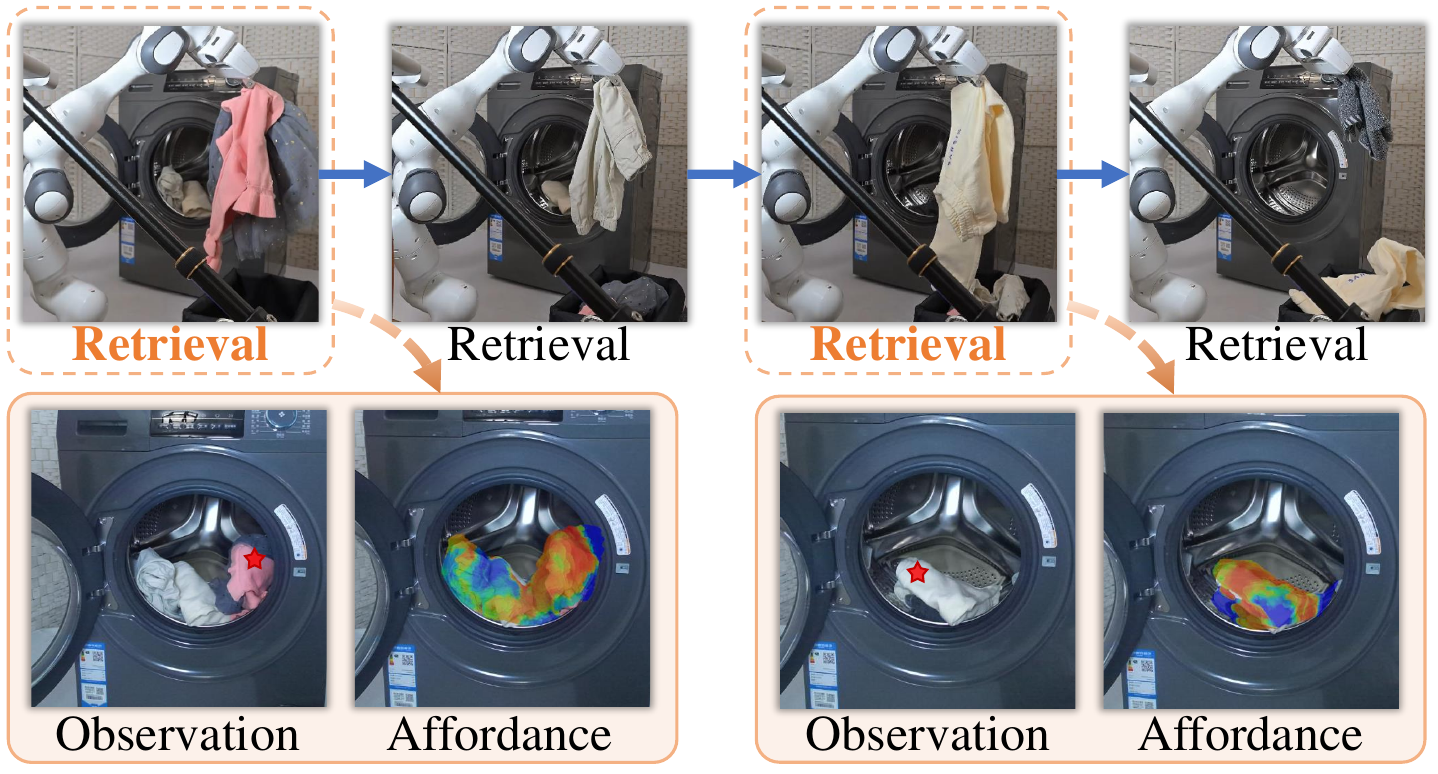}
  \end{center}
  \vspace{-5mm}\caption{\textbf{Washing Machine} Retrieval Sequence (without adaptation).}
  \vspace{-4mm}
\label{fig:realworld_wm}
\end{figure}

\begin{figure}[htbp]
  \begin{center}
   \includegraphics[width=1.0\linewidth]{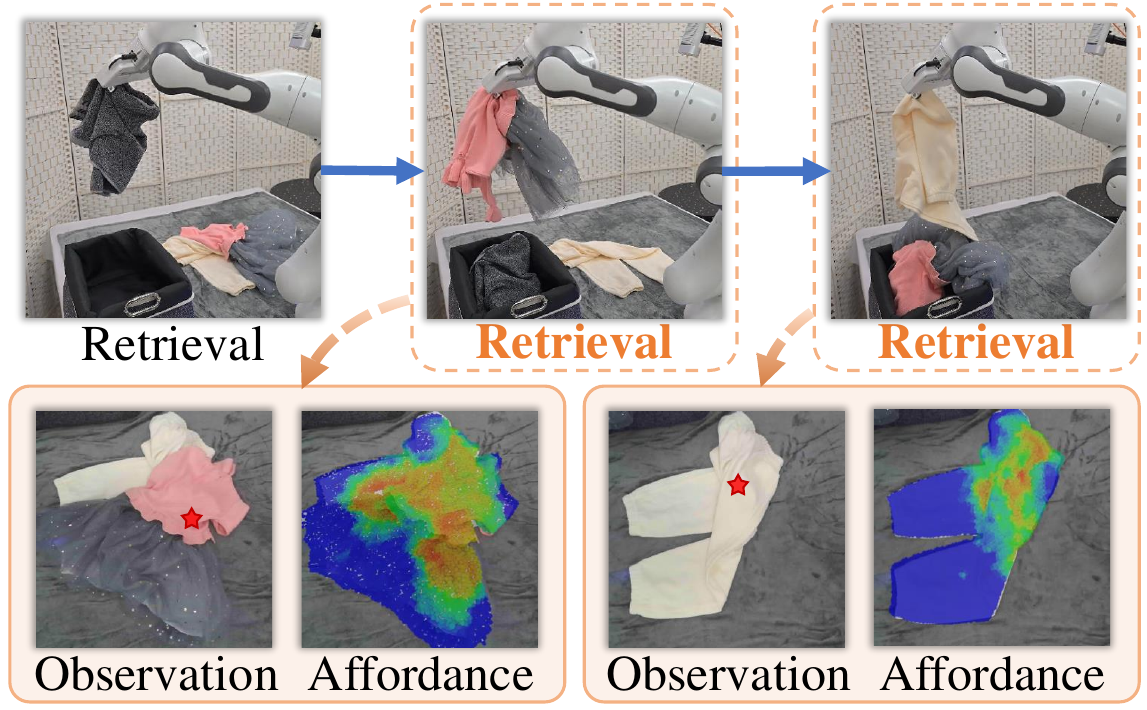}
  \end{center}
  \vspace{-5mm}\caption{\textbf{Sofa} Retrieval Sequence (without adaptation).}
  \vspace{-4mm}
\label{fig:realworld_sofa}
\end{figure}

\begin{figure}[htbp]
  \begin{center}
   \includegraphics[width=1.0\linewidth]{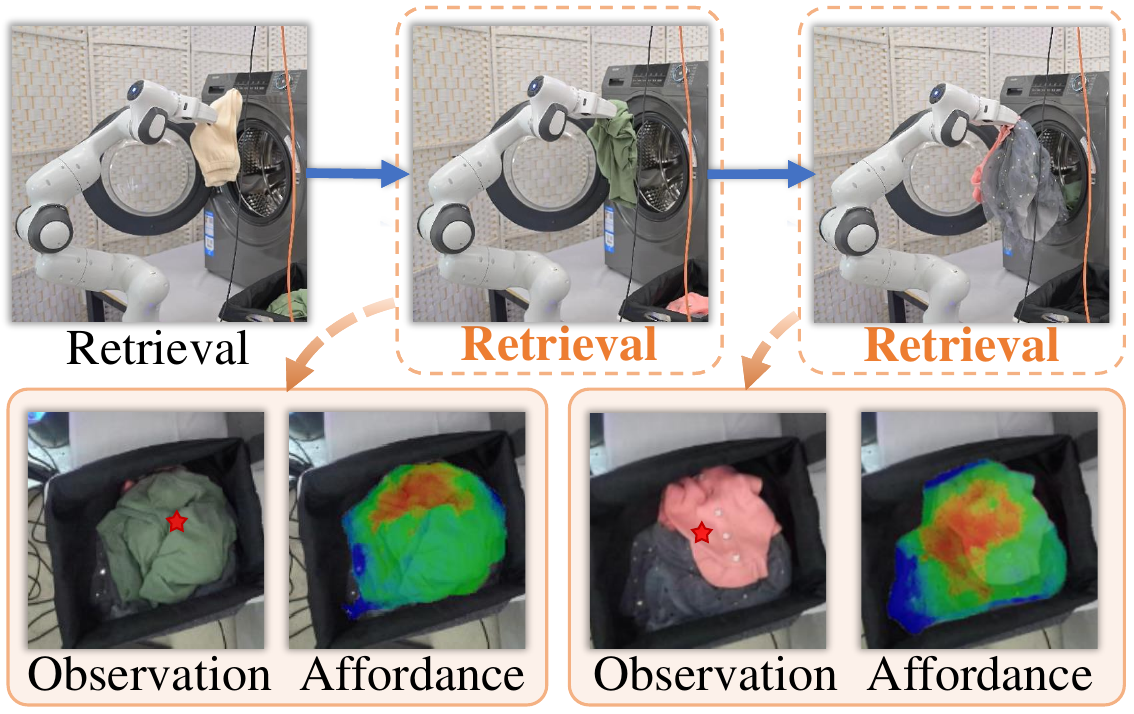}
  \end{center}
  \vspace{-5mm}\caption{\textbf{Basket} Retrieval Sequence (without adaptation).}
  \vspace{-3mm}
\label{fig:realworld_basket}
\end{figure}

We show the experimental results of the whole-process retrieval in the real-world washing machine scenario, real-world sofa scenario and real-world laundry basket scenario in Figure~\ref{fig:realworld_wm},~\ref{fig:realworld_sofa} and~\ref{fig:realworld_basket} respectively.

 Almost all retrieval operations are aware of the target garment's structure (the robot tends to grasp the middle part rather than the corners, even in scenarios with complex garment entanglements and severe occlusion) and the interrelation between garments (garments piled on the bottom or back generally do not produce highlights). Moreover, our affordance can also produce multi-modal output, in other words, when multiple pieces of garments can be retrieved, multiple highlights appear, and they are all reasonable.

\begin{figure}[htbp]
  \begin{center}
   \includegraphics[width=1.0\linewidth]{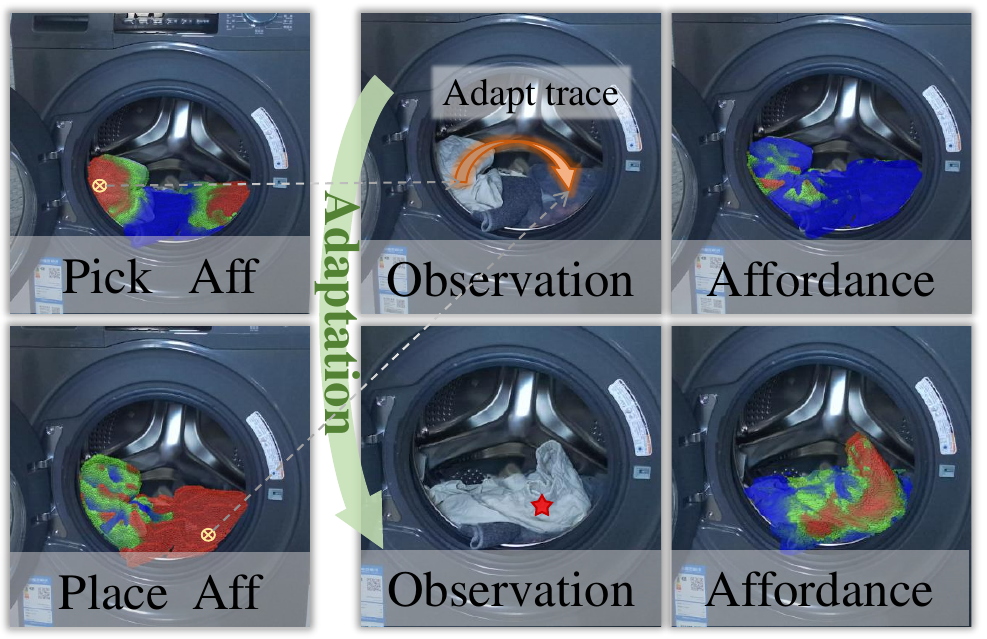}
  \end{center}
  \vspace{-5mm}\caption{\textbf {Washing Machine} Adaptation.}
  \vspace{-5mm}
\label{fig:adapt_wm}
\end{figure}

\begin{figure}[htbp]
  \begin{center}
   \includegraphics[width=1.0\linewidth]{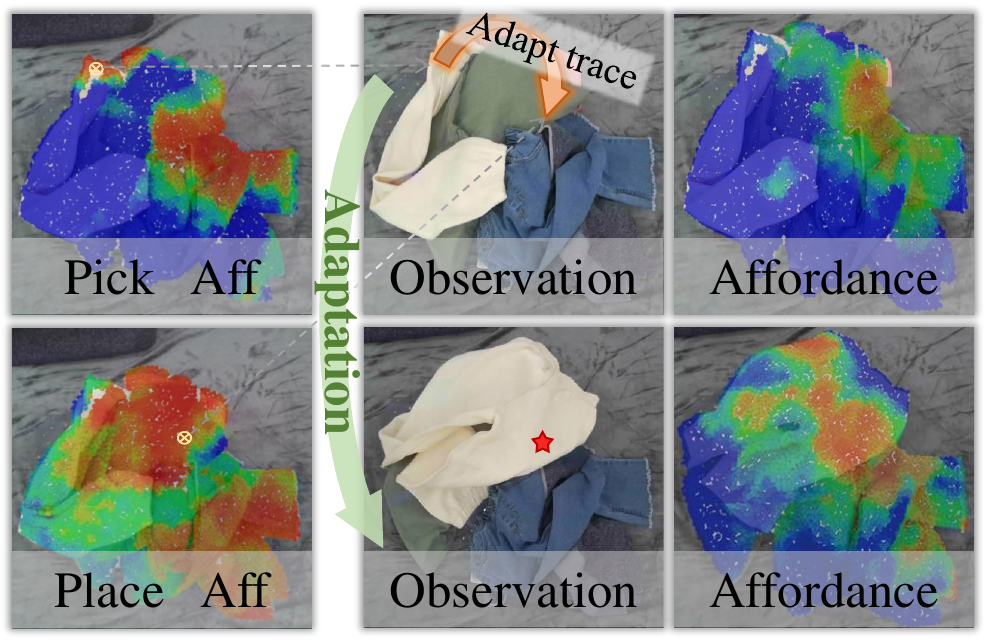}
  \end{center}
  \vspace{-5mm}\caption{\textbf {Sofa} Adaptation.}
  \vspace{-4mm}
\label{fig:adapt_sofa}
\end{figure}

\begin{figure}[htbp]
  \begin{center}
   \includegraphics[width=1.0\linewidth]{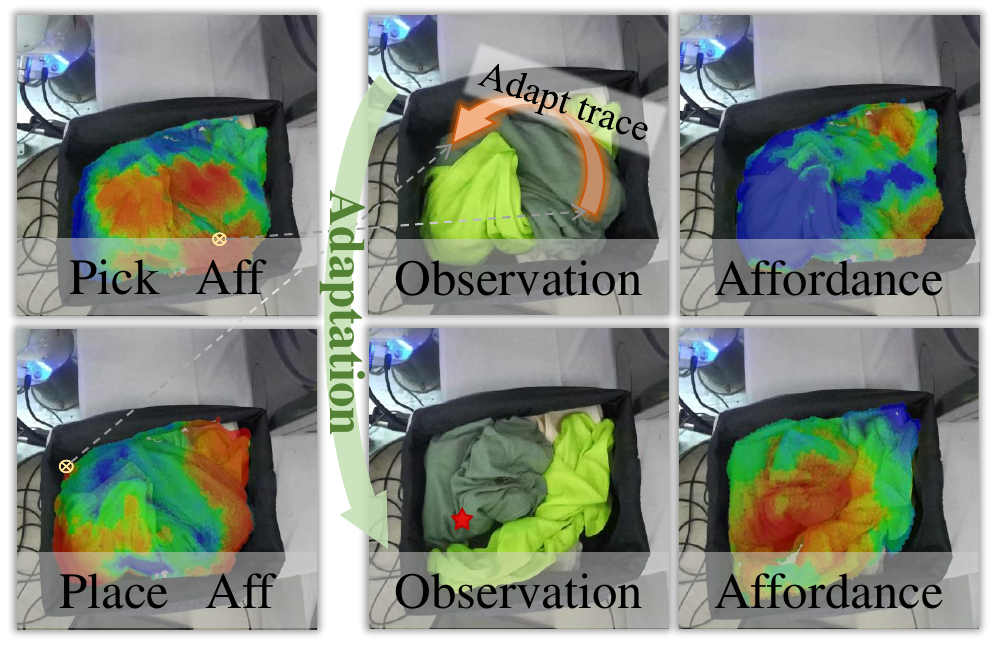}
  \end{center}
  \vspace{-5mm}\caption{\textbf {Basket} Adaptation.}
  \vspace{-5mm}
\label{fig:adapt_basket}
\end{figure}

We also tested our adaptation module in real-world scenarios. As shown in Figure~\ref{fig:adapt_wm},~\ref{fig:adapt_sofa} and~\ref{fig:adapt_basket}, when garments are severely tangled, the corresponding retrieval affordance appears poor. At this time, the model tends to execute an adaptation operation and find reasonable pick point and place point to adapt, and thus garments plausible for manipulation will exist.

\section{Offline training details}
We employed a random strategy to collect offline data, which enables us to select different but plausible points in similar observations, thereby capturing the multi-modal action distributions. Success rates of our offline trained models are \textbf{0.678}, \textbf{0.792}, \textbf{0.682} in 3 scenarios, consistently outperforming baselines (Tab.2 in main paper).
\section{Details of adaptation rounds}
Tab.~\ref{tab:adaptation} shows the relation of adaptation rounds and success rates.
With Tab.1 in main paper, we found up to 3 rounds of our proposed adaptation (instead of random adaptations) lead to plausible clutter states and make the success rate converge.

\begin{table}[!ht]
  \centering
  \begin{tabular}{@{}lccccc@{}}
    \toprule
    Rounds    & 3-rand & 0 & 1 & 2 & 3\\
    \midrule
    Success Rate & 0.719 & 0.712 & 0.782 & 0.803 & 0.805\\

    \bottomrule
  \end{tabular}
  \caption{\textbf{Success Rate on Different Adaptation Rounds.} 3-rand denotes 3 rounds of random adaptations.}
  \label{tab:adaptation}
\end{table}
\section{Why only raw PC without RGB?}
We agree using color as additional info can better distinguish scenes with very similar point cloud. However, there is a significant gap in color information between simulation and reality, particularly in low-light scenes like washing machine. 

\begin{figure}[!ht]
    \centering
    \begin{minipage}{\linewidth}
    \vspace{-3mm}\includegraphics[width=\textwidth]{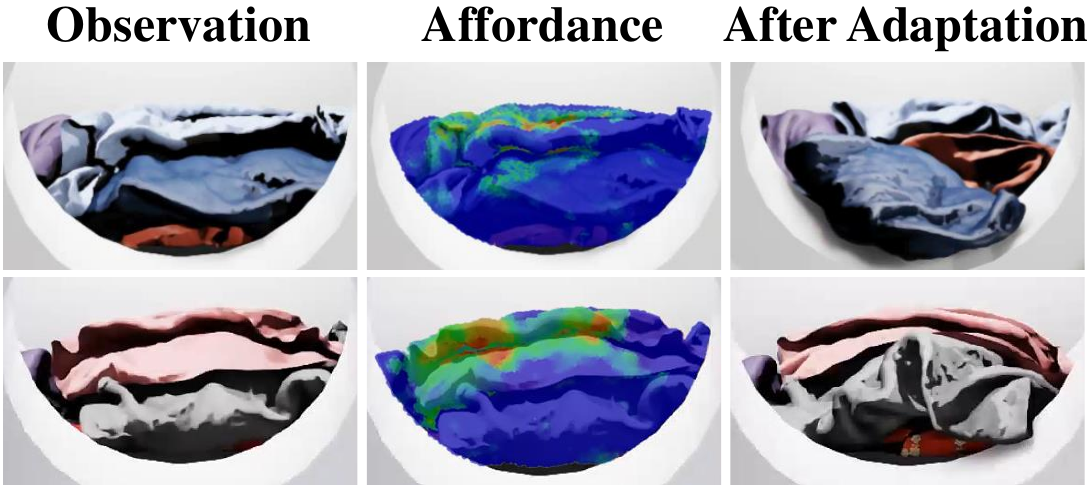}
    \vspace{-7mm}
    \caption{\textbf{Affordance and adaptation} of similar point cloud.}
    \vspace{-3mm}
    \label{fig:sim}
    \end{minipage}
\end{figure}

For two clutters with similar point clouds, the adaptation will help distinguish the clutters (Fig.~\ref{fig:sim}).

\begin{figure}[htbp]
  \vspace{-3mm}

  \begin{center}
   \includegraphics[width=0.9\linewidth]{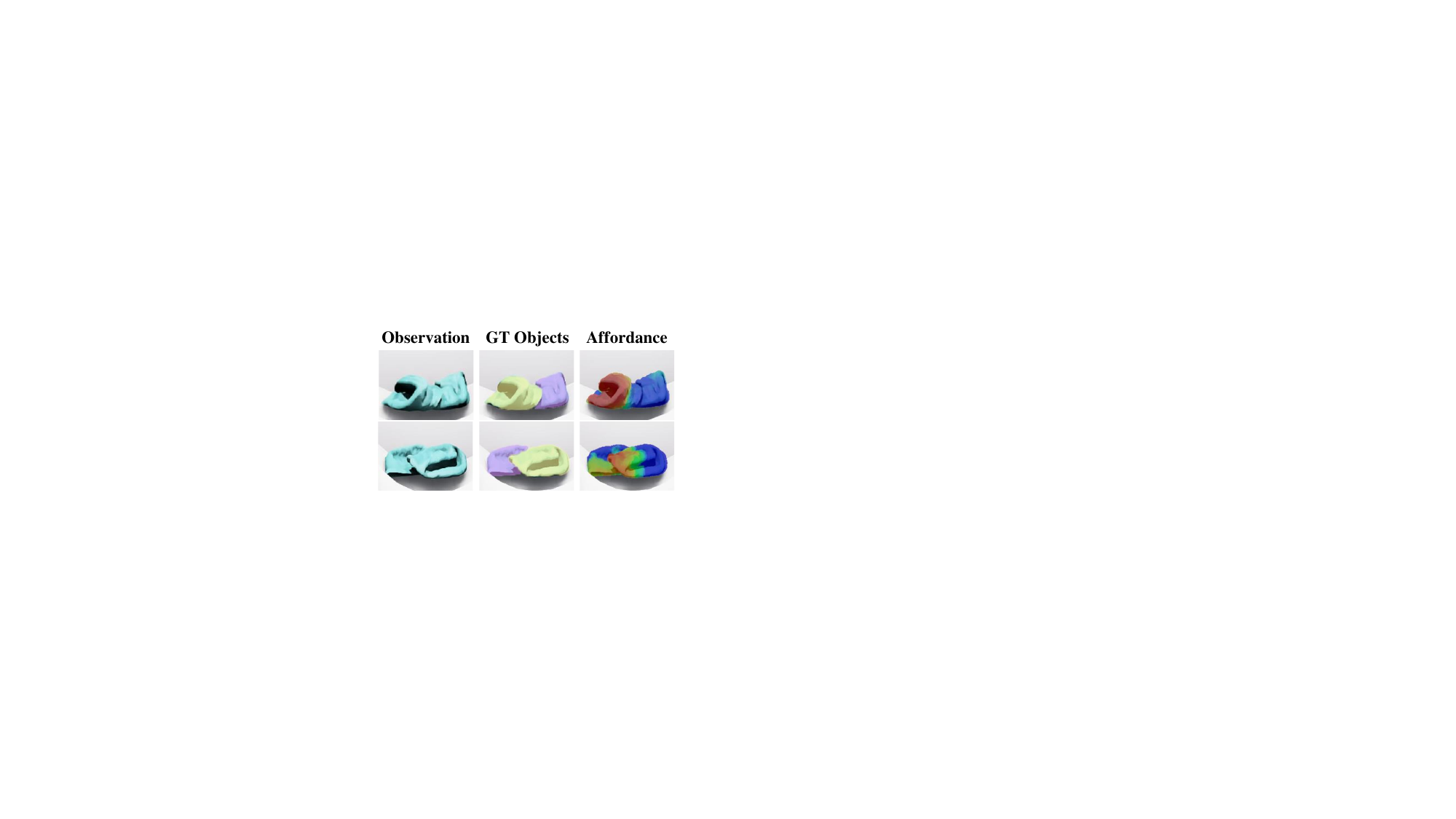}
  \end{center}
  \vspace{-5mm}\caption{\textbf{Point cloud} can distinguish similar-colored garments.}
  \vspace{-5mm}
\label{fig:pc}
\end{figure}

Point cloud (depth) is sensitive to wrinkles and spatial relations between garments with similar colors (Fig.~\ref{fig:pc}), enabling our method to effectively handle most clutters.

\section{Generalization to novel clutters}
\begin{figure}[!ht]
  \vspace{-2mm}

  \begin{center}
   \includegraphics[width=0.8\linewidth]{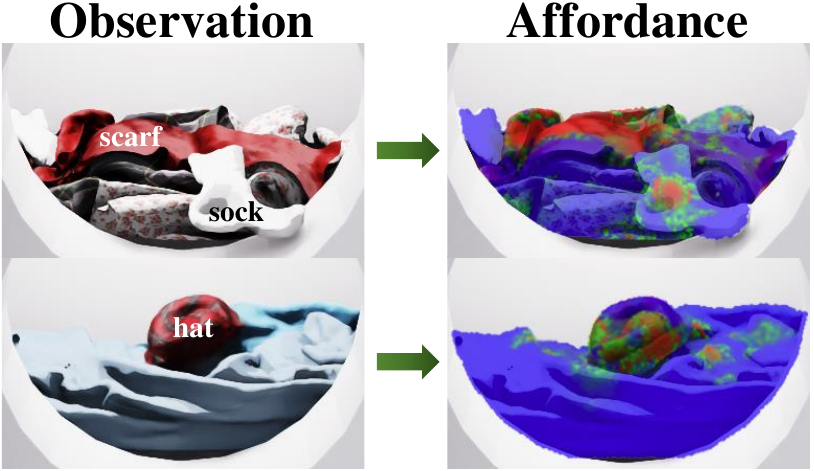}
  \end{center}
  \vspace{-5mm}\caption{Affordance on novel categories (scarf, sock and hat).}
  \vspace{-2mm}
\label{fig:novel}
\end{figure}

Each clutter is specific due to various garment states.
Besides, manipulation success rates of clutters with seen shapes, novel shapes in seen categories, and novel categories are \textbf{0.805}, \textbf{0.754} and \textbf{0.725} respectively.
Fig.~\ref{fig:novel}
shows affordance predictions in clutters with novel garment categories.

\section{Limitations}
For the simulation limitation, some extreme cases like knots between garments, cannot be simulated.
For such difficult cases,
manipulation requires more dexterous actions with 2 robots and even dexterous hands, instead of only parallel gripper's retrieving.
Other garment configurations and correlations (e.g., two garments are entangled) are possible.

\end{document}